\documentclass{article} 
\usepackage[accepted]{icml2024}

\usepackage{url}

\usepackage{graphicx}
\usepackage{booktabs}
\usepackage{caption}
\usepackage{subfig}
\usepackage{amsmath}
\usepackage{amssymb}
\usepackage{enumitem}
\usepackage{lscape}
\usepackage{multirow}
\usepackage{tablefootnote}
\usepackage{caption}

\icmltitlerunning{Binning as a Pretext Task: Improving Self-Supervised Learning in Tabular Domains}

\begin{document}

\twocolumn[
\icmltitle{
Binning as a Pretext Task: \\ Improving Self-Supervised Learning in Tabular Domains
}


\begin{icmlauthorlist}
\icmlauthor{Kyungeun Lee}{lgai}
\icmlauthor{Ye Seul Sim}{lgai}
\icmlauthor{Hye-Seung Cho}{lgai}
\icmlauthor{Moonjung Eo}{lgai}
\icmlauthor{Suhee Yoon}{lgai}
\icmlauthor{Sanghyu Yoon}{lgai}
\icmlauthor{Woohyung Lim}{lgai}
\end{icmlauthorlist}

\icmlaffiliation{lgai}{LG AI Research, Seoul, Repulic of Korea}

\icmlcorrespondingauthor{Woohyung Lim}{w.lim@lgresearch.ai}

\icmlkeywords{Machine Learning, ICML, Tabular, Tabular representation learning, Binning, Tabular learning}

\vskip 0.3in
]

\printAffiliationsAndNotice{\icmlEqualContribution}

\begin{abstract}
The ability of deep networks to learn superior representations hinges on leveraging the proper inductive biases, considering the inherent properties of datasets. In tabular domains, it is critical to effectively handle heterogeneous features (both categorical and numerical) in a unified manner and to grasp irregular functions like piecewise constant functions.
To address the challenges in the self-supervised learning framework, we propose a novel pretext task based on the classical \textit{binning} method. The idea is straightforward: \textit{reconstructing the bin indices} (either orders or classes) rather than the original values. This pretext task provides the encoder with an inductive bias to capture the irregular dependencies, mapping from continuous inputs to discretized bins, and mitigates the feature heterogeneity by setting all features to have category-type targets.
Our empirical investigations ascertain several advantages of binning: capturing the irregular function, compatibility with encoder architecture and additional modifications, standardizing all features into equal sets, grouping similar values within a feature, and providing ordering information. Comprehensive evaluations across diverse tabular datasets corroborate that our method consistently improves tabular representation learning performance for a wide range of downstream tasks. The codes are available in \url{https://github.com/kyungeun-lee/tabularbinning}.
\end{abstract}

\section{Introduction}
\label{sec:intro}

Tabular datasets are ubiquitous across diverse applications from financial markets and healthcare diagnostics to e-commerce personalization and manufacturing process automation. 
These datasets are structured with rows representing individual samples and columns representing heterogeneous features—a combination of categorical and numerical features—and they serve as the foundation for myriad analyses. 
Despite the wide applicability of tabular data, research into leveraging deep networks to harness the inherent properties of such datasets is still in its nascent stage.
In contrast, tree-based machine learning algorithms like XGBoost~\citep{chen2016xgboost} and CatBoost~\citep{prokhorenkova2018catboost} have consistently demonstrated prowess in discerning the nuances of tabular domains, outperforming deep networks even those with a larger model capacity and specialized modules~\citep{arik2021tabnet,gorishniy2021revisiting,grinsztajn2022tree,rubachev2022revisiting}. 
The consistent edge held by tree models fuels the exploration of how their advantageous biases can be adapted for deep networks.

Recently, the quest to boost the performance of deep networks on tabular data has gained momentum. A fundamental challenge is the inherent heterogeneity of tabular datasets, encompassing both categorical and numerical features~\citep{popov2019node,borisov2022deep,yan2023t2g}. 
To mitigate the feature discrepancies in deep networks, 
previous studies proposed using an additional module like a feature tokenizer~\citep{gorishniy2021revisiting} and an abstract layer~\citep{chen2022danets}.
Concurrently, some research has explored ways to infuse the proven strengths of tree-based models into deep networks. 
For instance, \citet{grinsztajn2022tree} observed that deep networks tend to prefer overly smooth solutions and struggle with modeling irregularities like piecewise constant functions, in contrast to the tree-based models. To address this challenge, \citet{gorishniy2022embeddings} introduced a novel approach combining piecewise linear encoding during preprocessing and periodic activation functions. Although these advancements have led to enhanced performance on several tabular data problems, they have predominantly been explored within a supervised learning framework, where they still fall short of outperforming simple tree-based methods~\cite{gorishniy2021revisiting,grinsztajn2022tree,mcelfresh2023neural}.

\begin{figure}[tb!]
    \centering
    \includegraphics[width=0.4\textwidth]{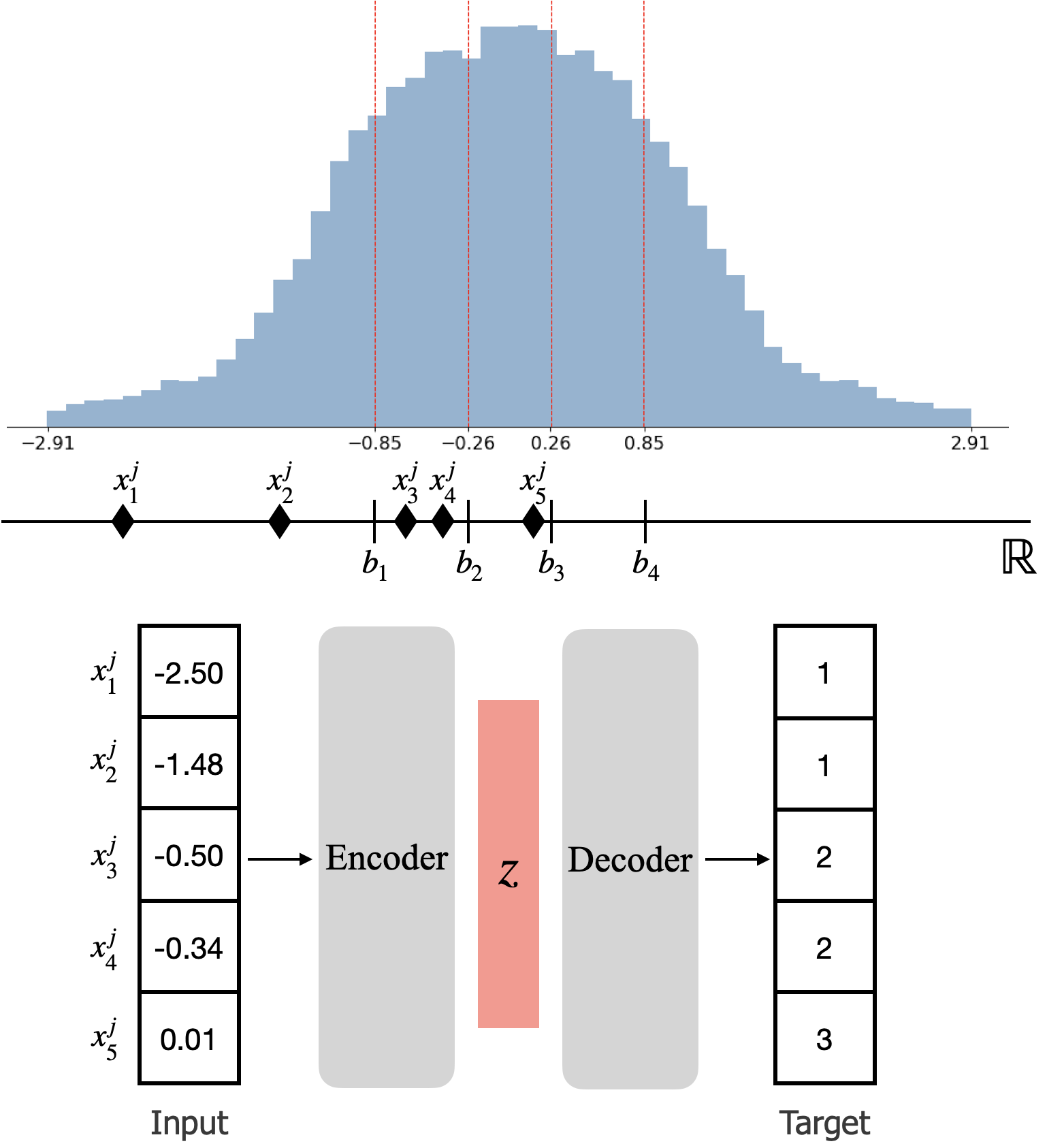}
    \caption{Binning as a pretext task. 
    Bins are determined based on the distribution of the training dataset for each feature. The inputs are passed into the encoder network, then the decoder network predicts the bin indices which can be {ordinal} when the pretext task is the regression or {nominal} when the pretext task is the classification.}
    \label{fig:fig1}
    \vspace{-0.5cm}
\end{figure}
%
In this study, we address the challenge of \textit{unsupervised} tabular deep learning where the tree-based methods are fundamentally inapplicable. To this end, we propose a novel pretext task based on the classical \textit{binning} method for auto-encoding-based self-supervised learning (SSL). 
Our approach is straightforward: \textit{reconstructing bin indices} rather than reconstructing the raw values, as illustrated in Figure~\ref{fig:fig1}.
Once numerical features are discretized into bins based on the quantiles of the training dataset, we optimize the encoder and decoder networks to accurately predict the bin indices given original inputs.
Despite its simplicity, binning as a pretext task offers several advantages for tabular deep learning. 
By setting the discretized bins as targets for the pretext task, we can employ the inductive bias of capturing the irregular functions and mitigating the discrepancy between features. 
The binning procedure allows grouping the nearby samples based on the distribution of the training dataset, 
so the learned representations should be robust to the minor errors that can yield spurious patterns. 
It also facilitates standardizing all features into equal sets, thereby preventing any uninformative features from dominating during SSL.  
Furthermore, our approach is compatible with any other modifications, including the choice of deep architectures and input transformation functions.

Based on the extensive experiments on 25 public datasets, we found that the binning task consistently improves the SSL performance on diverse downstream tasks, even though we simply changed the targets during SSL from the continuous to the discretized bins.
%
%
Finally, we found that the binning task can be not only an effective objective function for fully unsupervised learning but also beneficial as the pretraining strategy to achieve state-of-the-art performance, surpassing both tree-based and other supervised deep learning methods across a wide range of tabular data problems.

Our main contributions can be summarized as follows. First, we suggest binning as a new pretext task for SSL in tabular domains, compatible with any modifications.
Second, we conduct extensive experiments on 25 public tabular datasets focusing on the various input transformation methods and SSL objectives. 
Finally, we consistently achieve the best performance both in unsupervised and supervised learning frameworks.
The codes are available in \url{https://github.com/kyungeun-lee/tabularbinning}.
\section{Related works}
\label{sec:relatedworks}

\paragraph{Tabular deep learning: }
In recent years, there has been a large number of deep learning research on a tabular domain: developing new deep architectures~\citep{popov2019node,badirli2020grownet,huang2020tabtransformer,wang2021dcn,arik2021tabnet,gorishniy2021revisiting,chen2022danets,hollmann2022tabpfn,zhu2023xtab,kotelnikov2023tabddpm,chen2023trompt}; or representing the heterogeneous nature of tabular features as the graphs~\citep{yan2023t2g,chen2023hytrel}; or adopting new activation function~\citep{gorishniy2022embeddings}.
Despite these advancements, ensembles of decision trees, such as GBDTs (Gradient Boosting Decision Trees), continue to serve as competitive baselines~\citep{arik2021tabnet,gorishniy2021revisiting,grinsztajn2022tree,rubachev2022revisiting,mcelfresh2023neural,beyazit2023inductive}.
In this paper, our goal is to suggest a new pretext task for self-supervised learning in tabular domains, so we focus on architectures directly inspired by classic deep models, in particular MLPs and FT-Transformers~\citep{gorishniy2021revisiting}, in addition to the state-of-the-art tabular deep learning model, such as T2G-Former~\citep{yan2023t2g}.

\vspace{-0.4cm}
\paragraph{Self-supervised learning in tabular domains: }
Self-supervised learning (SSL) aims to learn desirable representations without making use of annotation information. 
Recently, contrastive learning and auto-encoding have been two major choices in the tabular domain.
Contrastive learning aims to model the similarity between two or more augmented views from the same sample, corresponding to the positive samples, and the dissimilarity between other samples, corresponding to the negative samples. 
\citet{bahri2021scarf,ucar2021subtab} have optimized contrastive loss after defining the positive and negative samples based on the data augmentation function, such as masking or cropping in feature dimension. 
Auto-encoding aims to reconstruct the original sample given its corrupted observation~\citep{vincent2008extracting}. 
Compared to contrastive learning, auto-encoders can handle a mix of data types which can be beneficial for tasks involving heterogeneous datasets, like tabular data.
\citet{yoon2020vime,huang2020tabtransformer,majmundar2022met} adopted the auto-encoding methods optimizing the reconstruction loss with or without the additional losses, such as corruption detection. 
In this study, we suggest a new SSL pretext task based on the auto-encoding approach.
\section{Backgrounds}

In this section, we delve into the auto-encoding-based self-supervised learning framework in tabular domains focusing on two factors: transformation methods to tabular inputs and the objective functions in the auto-encoding-based SSL framework.

\vspace{-0.2cm}

\begin{figure}[tb!]
    \centering
    \subfloat[Replacing value = Constant]{
    \label{fig:maskingconst}
    \includegraphics[width=0.85\columnwidth]{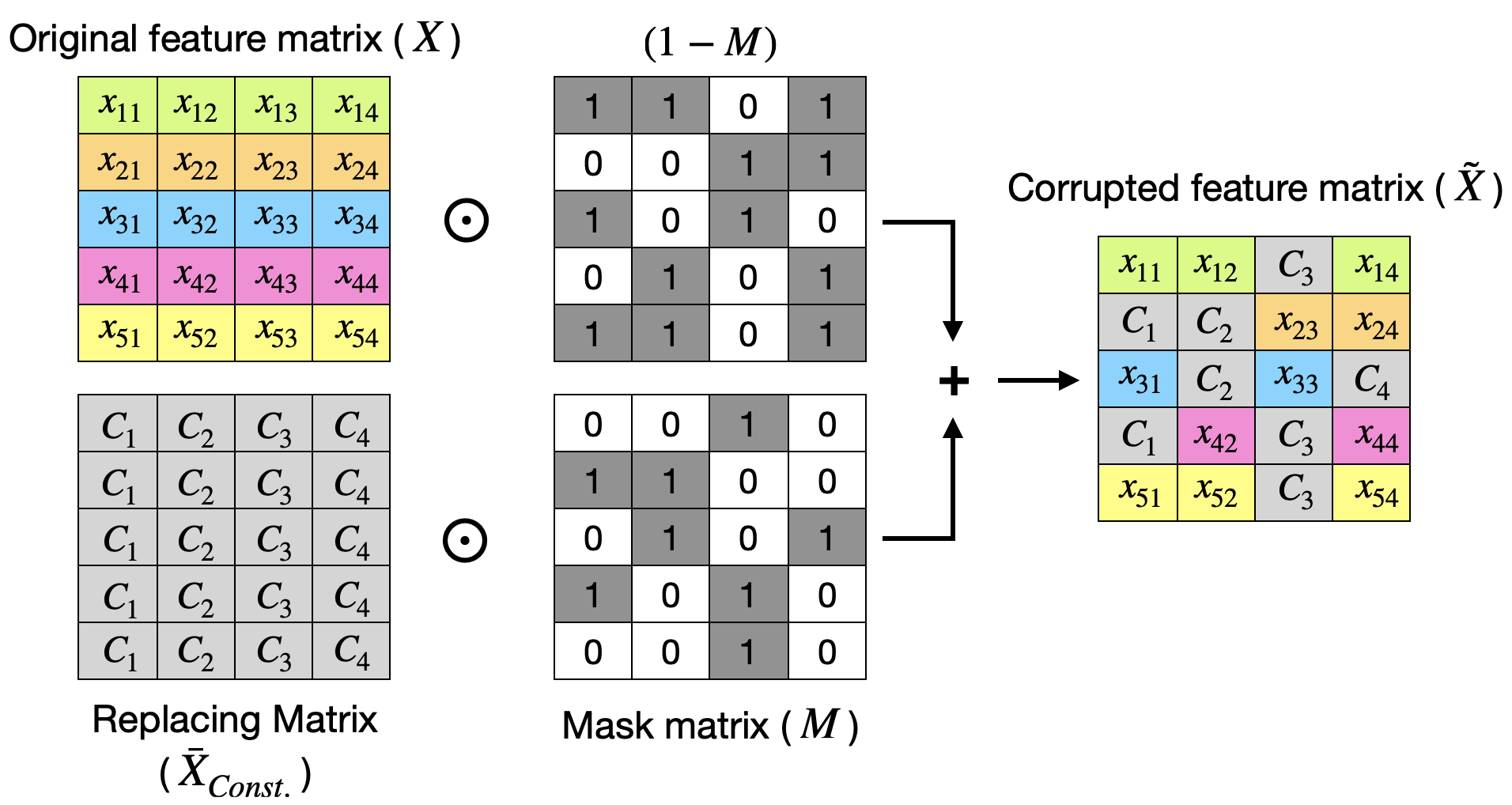}} \\
    \subfloat[Replacing value = Random]{
    \label{fig:maskingrand}
    \includegraphics[width=0.85\columnwidth]{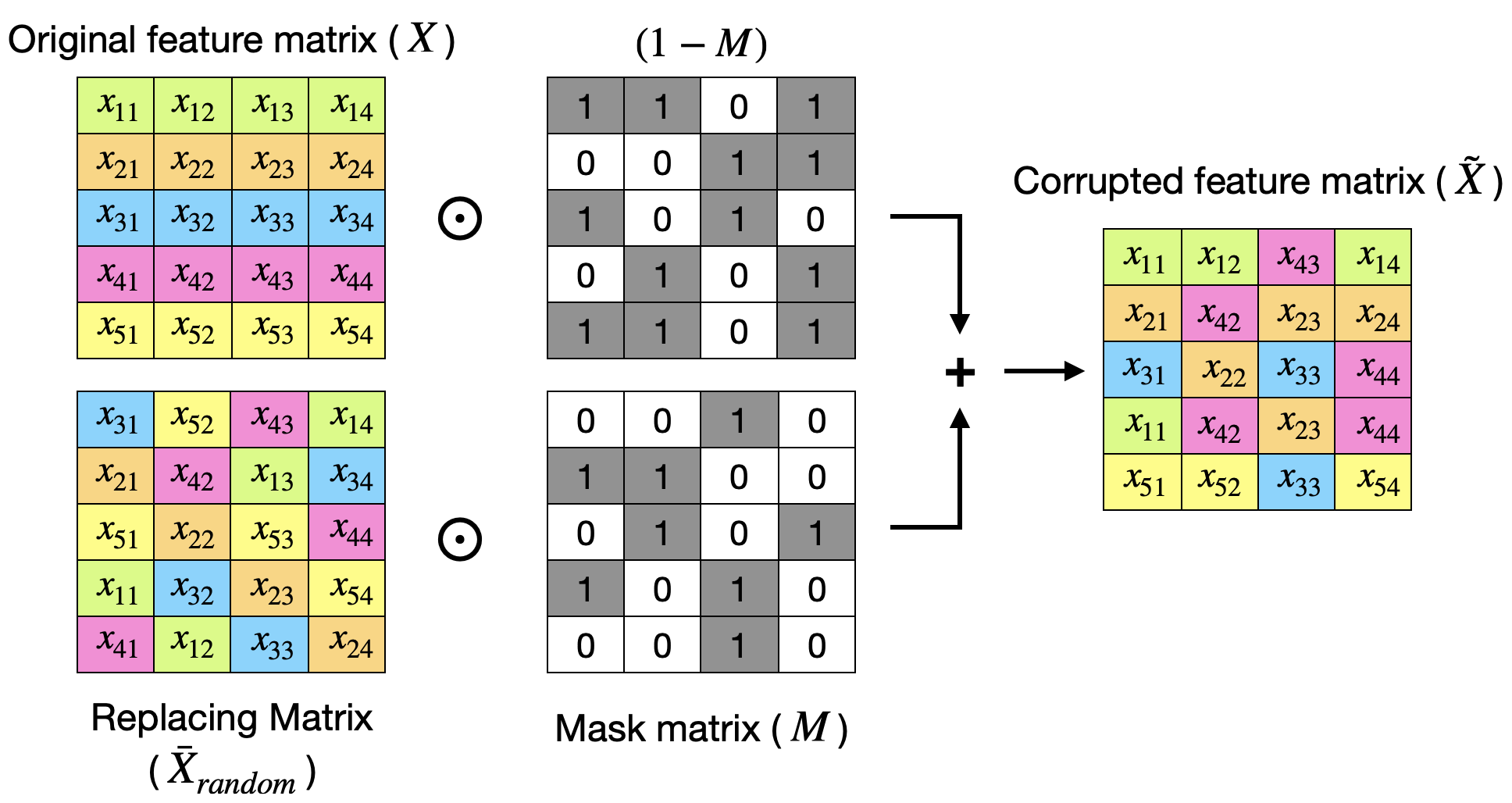}}
    \caption{An illustration of two methods to generate the replacing vectors for masked features.}
    \label{fig:shuffling}
    \vspace{-0.3cm}
\end{figure}

\paragraph{Input transformation: } 
To ensure the encoder network does not simply learn an identity function, we employ transformation functions on the input that preserve the label-related information. 
For tabular datasets, only a few transformation functions have been suggested like masking~\citep{yoon2020vime,ucar2021subtab,majmundar2022met} as illustrated in Figure~\ref{fig:shuffling} because all individual values can play a key role in determining the semantics and small changes can affect the context.
Given a sample $x_i\in\mathbb{R}^d$ in dataset $\mathcal{D}$ where $d$ is the number of features, $i\in [1, N]$, and $N$ is the batch size, we randomly generate the masking vector $m_i$ with the same size of $x_i$. 
Each element of the masking vector $m_{i}$ is independently sampled from a Bernoulli distribution with probability $p_m\in [0, 1]$. 
To replace the masked values, the replacing vector $\bar{x}_i$ should be defined. In this study, we utilize two methods suggested in the previous studies~\citep{yoon2020vime,ucar2021subtab,majmundar2022met}.
\begin{itemize}[leftmargin=*]
    \vspace{-0.2cm}
    \item Constant (Figure~\ref{fig:maskingconst}): $\bar{x}_{i,k}$ is set as the pre-determined constant value for all $i$. In this study, we use the average for each feature $k$ in the training dataset.
    \vspace{-0.2cm}
    \item Random (Figure~\ref{fig:maskingrand}): $\bar{x}_{i,k}$ is sampled from the other in-batch samples for a given feature. In other words, to replace the $k$-th feature of the $i$-th sample in the batch, we use the $k$-th feature of the $i'$-th sample in the same batch, and $i'$ is sampled from the uniform distribution $\mathcal{U}\left(\frac{1}{N}\right)$. 
    \vspace{-0.2cm}
\end{itemize}

Finally, the corrupted sample $\tilde{x}_i$ is formulated as $\tilde{x}_i = (\textbf{1}-m_i) \odot x_i + m_i \odot \bar{x}_i$ where $\textbf{1}$ is all-ones vector with the same size of $x_i$.  
The transformation procedure is stochastic and it provides randomness during training. When $p_m=0$, $m_i$ becomes the zero matrix, and the uncorrupted input $\tilde{x}_i=x_i$ is used for training.

\vspace{-0.2cm}
\paragraph{SSL objectives: } 

Following the convention of SSL, the encoder $f_e$ first transforms the corrupted sample $\tilde{x}_i$ to a representation $z_i$, then the decoder $f_d$ will be introduced to learn the informative representation by optimizing the unsupervised loss $\mathcal{L}$. 
We can leverage which representation should be learned by introducing the specific pretext task.
As a baseline, we consider two pretext tasks used in \citet{yoon2020vime,huang2020tabtransformer,majmundar2022met}.
\begin{itemize}[leftmargin=*]
    \vspace{-0.2cm}
    \item Reconstructing the original values: 
    One common approach is to reconstruct uncorrupted samples from their corrupted counterparts~\citep{vincent2008extracting}. In this setup, the encoder attempts to impute the masked features by leveraging the correlations present in the non-masked features. The learned representations will involve the semantic-level information that is invariant to corruption. 
    To this end, the decoder network is defined as $f_d^\text{recon}: Z\rightarrow \hat{X}$, and the corresponding loss is formulated as 
    $\mathcal{L}_\text{ValueRecon} := \frac{1}{N}\sum_{i=1}^{N}||x_i-f_d^\text{recon}(z_i)||_2^2$.
    \vspace{-0.2cm}
    \item Detecting the masked features: 
    An auxiliary task that can facilitate the pretext task of reconstruction is predicting which features have been masked during the corruption process of the input sample~\citep{yoon2020vime}. 
    In this setup, the encoder attempts to leverage the inconsistency between feature values to identify the masked features, resulting in learned representations that capture abnormal patterns for a given input.
    Specifically, the method employs a binary cross-entropy loss, which can be formulated as $\mathcal{L}_\text{MaskXent} := -\frac{1}{N}\sum_{i=1}^{N} m_i\log{f_d^\text{mask}(z_i)}+ (\textbf{1}-m_i)\log{(\textbf{1}-f_d^\text{mask}(z_i))}$ where the decoder network is defined as $f_d^\text{mask}: Z\rightarrow \hat{M}$.
    \vspace{-0.2cm}
\end{itemize}

We can optimize several loss functions simultaneously if we train several decoders that utilize $z$ as the inputs. For example, \citet{yoon2020vime} utilized the weighted sum of $\mathcal{L}_\text{ValueRecon}$ and $\mathcal{L}_\text{MaskXent}$. 

\section{Methods: Binning as a Pretext Task for Tabular SSL}
\label{sec:methods}

\begin{figure}[tb!]
    \centering
    \includegraphics[width=0.49\textwidth]{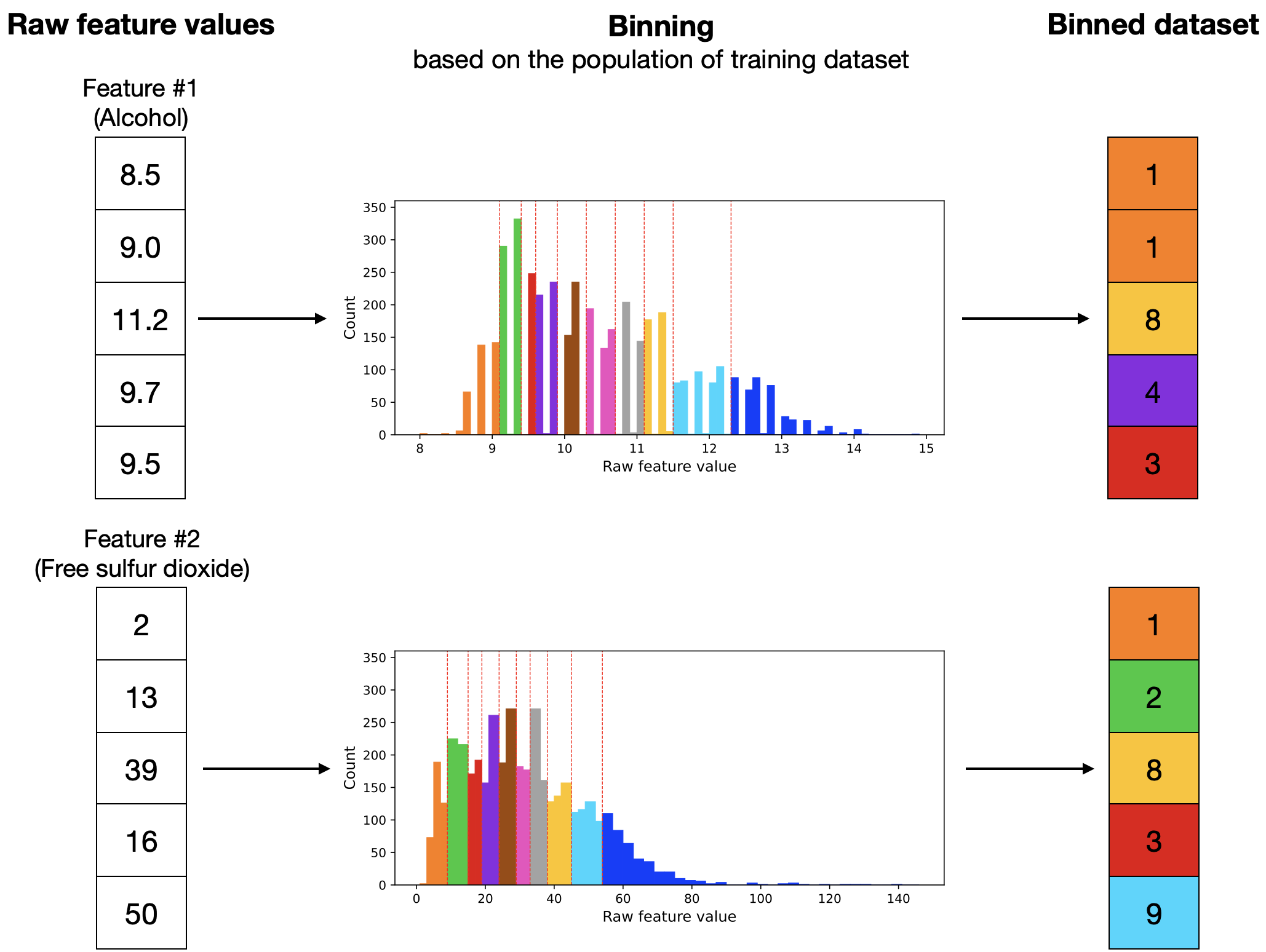}   
    \vspace{-0.3cm}
    \caption{An example of binning (Dataset: Wine Quality~\citep{cortez2009modeling}). 
    In the example, we set $T$ as 10. For each feature, we implement the binning to include the same number of observations based on the training dataset. Finally, we use the binning indices as the targets for auto-encoding-based SSL. 
    When we regard the bin indices as the classes without order information, the binning indices are converted into the one-hot vectors.
    }
    \vspace{-0.3cm}
    \label{fig:binning}
\end{figure}

Binning is a classical data preprocessing technique that quantizes a given numerical feature $x^j\in\mathbb{R}^{|\mathcal{D}|}$ into $T$ discrete intervals, known as \textit{bins} $B_t^j=[b_{t-1}^j, b_t^j)$ where $t\in [1, T]$ and $b_t^j\in\mathbb{R}$ is the bin boundaries. 
Binning is effective in transforming continuous features into discrete ones, mitigating minor errors in datasets like noise and outliers, and making the data distribution more manageable~\citep{dougherty1995supervised,han2022data}.

In this study, we implement binning to establish targets for auto-encoding-based SSL. We anticipate the representations will be robust to the minor input variation in the same bins. Also, the deep networks can capture the irregularities akin to the decision-making process of tree-based models, which assign discrete leaves to each continuous sample because the pretext task corresponds to mapping continuous inputs to discretized bins.
In addition, the binning approach helps mitigate feature heterogeneity by treating the targets for all features as the same category type during SSL. 

\citet{muller2021transformers} proposed a similar approach in the context of Bayesian inference, addressing the well-known challenge that neural networks face in modeling continuous distributions. To overcome this issue, they pivoted towards utilizing discretized continuous distributions for accurately modeling posterior probability distributions. Their research revealed that integrating discretization into the objective function of deep networks not only enhances the effectiveness of the training process but is also theoretically established as a versatile technique capable of modeling any distribution. This methodology highlights the substantial potential of binning in augmenting the capabilities of neural networks.

The binning procedure is described in Figure~\ref{fig:binning}.
We first determine the number of bins $T$ as the design parameter. Then, we split the value range into the disjoint set of $T$ intervals, 
$\left\{ B_1^j, \dots, B_T^j \right\}$, 
considering the number of observations in the training dataset $\mathcal{D}_\text{train}$ for each $j$-th feature $x^j$.
Specifically, the bin boundaries $b_t^j$ are determined according to the quantiles of $\frac{t}{T}$.
(Alternative binning strategies are also discussed in Supplementary~\ref{sup_sec:binmethod}.)
When the number of unique values for $x^j$ in the training dataset is less than $T$, each distinct value is assigned its own bin. 
Finally, we place each numerical feature $x_i^j$ into the bin $B_t^j$,
and we substitute the original values with the corresponding bin indices $t_i^j\in [1, T]$. 
Thus, we use the grouped ranks (or classes) instead of the raw values. 
We call the binned dataset as $X_\text{Bin}$.

The bin index of $i$-th sample and $j$-th feature, $t_i^j$, can be expressed as ordinal values or nominal classes. 
When we utilize the bin indices as ordinal values, we set the pretext task as reconstructing the bin indices based on the continuous inputs, and the corresponding \textit{BinRecon loss} is defined as 
\begin{align*}
    \vspace{-0.15cm}
    \mathcal{L}_\text{BinRecon} := 
    \frac{1}{N}\sum_{i=1}^N \left\| t_i - f_d^\text{BinRecon}(z_i) \right\|_2^2 \\
    \text{ where } f_d^\text{BinRecon}: Z\rightarrow \hat{X}_\text{Bin}.
    \vspace{-0.07cm}
\end{align*}
When we utilize the bin indices as nominal classes, 
we convert the bin index $t^j_i$ into the one-hot vector $\textbf{u}_i^j=\left[ u_1, u_2, \dots, u_T \right]$ where $u_v=1$ when $v=t_i^j$ and $u_v=0$ otherwise. 
Then, we set the pretext task as predicting the bin indices as classes by optimizing the \textit{BinXent loss}, defined as the multi-class cross-entropy loss for each feature.
\begin{align*}
    \vspace{-0.15cm}
    \mathcal{L}_\text{BinXent}:=
    -\frac{1}{Nd}\sum_{i=1}^{N} \sum_{j=1}^{d} \textbf{u}_i^j\log{f_d^\text{BinXent}(z_i^j)}
    \vspace{-0.07cm}
\end{align*}
In this case, the predictions for each sample should be in $\mathbb{R}^{d\times T}$. As a simple implementation, we add the 1x1 convolutional layer at the end of $f_d^\text{BinXent}(\cdot): Z\rightarrow \hat{U}$ where $U\in\mathbb{R}^{N\times d\times T}$ represents the one-hot encoded binned dataset.

We outline the benefits of utilizing the binning task in SSL as follows.
Empirical evidence on how each item is advantageous for tabular data problems will be provided in subsequent sections.

\vspace{-0.2cm}
\begin{itemize}[leftmargin=*]
    \item \textit{Capturing the irregular function}: 
    We explicitly make deep networks learn the function that maps from continuous inputs to discrete targets during SSL. It effectively provides beneficial inductive bias for tabular learning and mitigates the discrepancy between heterogeneous features.
    (Section~\ref{subsec:results2}, \ref{subsec:discussion_bininfo}, \ref{subsec:discussion_vis})
    \item \textit{Compatibility with other modifications}: 
    The binning task is agnostic to modifications such as changes in encoder architecture, input transformation functions, and additional objectives. Thus, it can be utilized independently or in conjunction with other options.
    (Section~\ref{subsec:results1}, \ref{subsec:results2})
    \item \textit{Standardizing all features into equal sets}\footnote{Detailed description is available in Supplementary~\ref{supsec:uninformative}.}: 
    After binning, all features lie on the uniform distribution with identical elements. Unlike the conventional normalization schemes, it largely simplifies the dataset to include only $T$ distinct values, and 
    this ensures all features become equal sets,
    thereby preventing any uninformative features from dominating during training.
    (Section~\ref{subsec:ablation})
    \item \textit{Grouping similar values in each feature}: 
    Binning clusters the nearby values in each feature and eliminates the other information except the bin index. Deep networks can identify nearby samples in a distribution as similar, independent of their magnitude. (Section~\ref{subsec:ablation})
    \item \textit{Ordering in BinRecon loss}: 
    BinRecon loss utilizes the grouped rank information only while eliminating the raw value information.
    This ensures that the encoder network learns the ordering information, regardless of the magnitude of the values.
    (Section~\ref{subsec:ablation})
\end{itemize}
\vspace{-0.2cm}

Overall, we implement SSL as follows. 
First, tabular inputs undergo a transformation that retains their semantic information.
Then, the encoder network $f_e$ takes the transformed input $\tilde{x}$ and produces the representation $z$, and the decoder network $f_d$ models the representation $z$ to the target $\hat{y}_\text{SSL}$ depending on the choice of pretext task. In this study, we consider four types of pretext tasks and the corresponding losses are ValueRecon, MaskXent, BinRecon, and BinXent. Once SSL is finished, the learned representations $z$ are evaluated based on linear probing.
\section{Experiments}
\label{sec:results}

In this section, we evaluate the effectiveness of binning as a pretext task across 25 public tabular datasets encompassing a range of data sizes and task types. Dataset details are provided in Supplementary~\ref{sup_sec:dataset}. 
For all datasets, we apply standardization for numerical features and labels for evaluating the regression tasks.

For the encoder network $f_e$, we experiment three types of deep networks: (1) MLPs, representing the simplest form of deep architecture; (2) FT-Transformer~\cite{gorishniy2021revisiting}, a simple adaptation of the Transformer architecture for the tabular domain; and (3) T2G-Former~\cite{yan2023t2g}, the state-of-the-art deep architecture for tabular data problems.
Note that a larger or more complex network does not guarantee better performance in tabular datasets~\citep{gorishniy2021revisiting,rubachev2022revisiting,grinsztajn2022tree,gorishniy2022embeddings,mcelfresh2023neural}.
To determine the depth and width of $f_e$ in the case of MLP, we identify the optimal configuration based on validation performance in the supervised setup, i.e., only the encoder with a linear head is trained with the supervised loss, ensuring the unsupervised nature of our framework. In the case of FT-Transformer and T2G-Former, we use the default setup from the original paper. For the decoder $f_d$, we always employ the MLP architecture as the same as the MLP-type encoder network. Consequently, all cases for each dataset have been trained on the same architecture and optimization setups. A detailed description is provided in Supplementary~\ref{sup_sec:implementation}.
For a given network and dataset, we also investigate the masking probability $p_m\in\left\{ 0.1, 0.2, 0.3, 0.4, 0.5, 0.6, 0.7, 0.8, 0.9 \right\}$ and the number of bins $T\in\left\{ 2, 5, 10, 20, 50, 100 \right\}$. Then, we found the optimal configuration based on validation performance on each downstream task. After SSL, we evaluate the representations based on linear probing 10 times with different random seeds, and an average is reported. 
We evaluate the representation quality based on accuracy for classification tasks and RMSE for regression tasks.
The full results with standard deviation are also available in Supplementary~\ref{sup_sec:full_results}. All experiments are conducted on a single NVIDIA GeForce RTX 3090. The codes are available in \url{https://github.com/kyungeun-lee/tabularbinning}.

\subsection{Comparison with the unsupervised methods: Linear evaluation results}
\label{subsec:results1}

\begin{table*}[tb!]
\caption{Linear evaluation results for various SSL methods when the encoder network is fixed as MLP. 
For each method, we also determine the performance rankings for each dataset, and the average ranks are also provided in the last column. Best cases for each dataset are marked in \textbf{bold}.}
\label{tab:ssl}
\centering
\subfloat[Binary classification (Metric: Accuracy)]{
    \label{tab:binclass}
    \resizebox{0.91\textwidth}{!}{
    \begin{tabular}{lllccccccccc}
    \toprule
    Masking & Replacing value & SSL Objective(s) & CH & HI & AD & BM & PH & OS & CS & PO & Average Rank \\\cmidrule(lr){1-3}\cmidrule(lr){4-11}\cmidrule(lr){12-12}
    FALSE & - & ValueRecon & 0.810 & 0.651 & 0.837 & 0.899 & 0.728 & 0.883 & 0.709 & 0.851 & 7.571 \\
    TRUE & Const. & MaskXent & 0.807 & 0.672 & 0.836 & 0.899 & 0.715 & 0.893 & 0.708 & 0.845 & 7.286 \\
    TRUE & Const. & ValueRecon & 0.810 & 0.653 & 0.839 & 0.900 & 0.734 & 0.884 & 0.718  & 0.849 & 6.429 \\
    TRUE & Const. & MaskXent+ValueRecon & 0.817 & 0.669 & 0.835 & 0.900 & 0.724 & 0.877 & 0.706 & 0.837 & 7.714 \\
    TRUE & Random & MaskXent & 0.814 & 0.681 & 0.843 & \textbf{0.901} & 0.710 & 0.883 & 0.706 & 0.853 & 5.429 \\
    TRUE & Random & ValueRecon & 0.811 & 0.661 & 0.838 & 0.898 & 0.736 & 0.885 & 0.714 & 0.842 & 7.143 \\
    TRUE & Random & MaskXent+ValueRecon & 0.804 & 0.647 & 0.826 & 0.899 & 0.715 & 0.879 & 0.713 & 0.861 & 8.571 \\\midrule
    FALSE & - & BinXent & 0.817 & 0.683 & 0.845 & \textbf{0.901} & 0.732 & 0.886 & \textbf{0.738} & 0.851 & 3.571 \\
    FALSE & - & BinRecon & \textbf{0.823} & \textbf{0.687} & 0.840 & 0.900 & \textbf{0.737} & 0.889 & 0.724 & \textbf{0.865} & \textbf{2.286} \\
    TRUE & Const. & BinRecon & 0.820 & 0.672 & 0.843 & 0.899 & 0.730 & \textbf{0.896} & 0.718 & 0.858 & 3.714 \\
    TRUE & Random & BinRecon & 0.819 & 0.682 & \textbf{0.846} & 0.898 & 0.735 & 0.894 & 0.718 & 0.858 & 3.571 \\\bottomrule          
    \end{tabular}
}} \\
\subfloat[Multiclass classification (Metric: Accuracy)]{
    \label{tab:multiclass}
    \resizebox{0.98\textwidth}{!}{
    \begin{tabular}{lllcccccccccc}
    \toprule
    Masking & Replacing value & SSL Objective(s) & CO & OT & GE & VO & WQ & AL & HE & MNIST & p-MNIST & Average Rank \\\cmidrule(lr){1-3}\cmidrule(lr){4-12}\cmidrule(lr){13-13}
    FALSE & - & ValueRecon & 0.769 & 0.776 & 0.527 & 0.619 & 0.568 & 0.931 & 0.353 & 0.965 & 0.928 & 6.333 \\
    TRUE & Const. & MaskXent & 0.784 & 0.777 & 0.518 & 0.545 & 0.547 & 0.909 & 0.341 & 0.793 & 0.554 & 9.333 \\
    TRUE & Const. & ValueRecon & 0.783 & 0.791 & 0.557 & 0.622 & 0.586 & 0.931 & 0.354 & 0.966 & 0.925 & 4.111 \\
    TRUE & Const. & MaskXent+ValueRecon & 0.750 & 0.774 & 0.519 & 0.610 & 0.571 & 0.931 & 0.360 & 0.941 & 0.907 & 7.444 \\
    TRUE & Random & MaskXent & 0.763 & 0.791 & 0.555 & 0.549 & 0.544 & 0.925 & 0.336 & 0.945 & 0.817 & 8.000 \\
    TRUE & Random & ValueRecon & 0.761 & 0.782 & 0.538 & 0.625 & 0.573 & 0.930 & 0.357 & 0.956 & 0.934 & 5.556 \\
    TRUE & Random & MaskXent+ValueRecon & 0.769 & 0.779 & 0.521 & 0.564 & 0.519 & 0.925 & 0.353 & 0.945 & 0.906 & 8.333 \\\midrule
    FALSE & - & BinXent & 0.742 & 0.781 & 0.517 & 0.600 & 0.565 & 0.903 & 0.354 & 0.956 & 0.908 & 8.333 \\
    FALSE & - & BinRecon & 0.784 & 0.783 & 0.544 & 0.625 & \textbf{0.592} & 0.935 & 0.357 & 0.964 & 0.950 & 3.556 \\
    TRUE & Const. & BinRecon & 0.812 & 0.792 & 0.559 & 0.647 & 0.581 & 0.943 & 0.359 & 0.974 & 0.964 & 2.222 \\
    TRUE & Random & BinRecon & \textbf{0.814} & \textbf{0.794} & \textbf{0.580} & \textbf{0.655} & 0.574 & \textbf{0.949} & \textbf{0.365} & \textbf{0.981} & \textbf{0.971} & \textbf{1.333} \\\bottomrule          
    \end{tabular}
    }
} \\
\subfloat[Regression (Metric: RMSE)]{
    \label{tab:regression}
    \resizebox{0.95\textwidth}{!}{
    \begin{tabular}{lllccccccccc}
    \toprule
    Masking & Replacing value & SSL Objective(s) & CA & HO & FI & MI & KI & CPU & DIA & EL & Average Rank \\\cmidrule(lr){1-3}\cmidrule(lr){4-11}\cmidrule(lr){12-12}
    FALSE & - & ValueRecon & 0.749 & 4.241 & 13900.720 & 0.784 & 0.163 & 3.876 & 1016.641 & 0.399 & 8.625 \\
    TRUE & Const. & MaskXent & 0.709 & 4.548 & 13473.750 & 0.788 & 0.185 & 4.475 & 1259.744 & 0.396 & 8.875 \\
    TRUE & Const. & ValueRecon & 0.693 & 4.086 & 13518.683 & 0.778 & 0.160 & 3.728 & $\phantom{0}$952.444 & 0.394 & 5.000 \\
    TRUE & Const. & MaskXent+ValueRecon & 0.700 & 4.157 & 13915.875 & 0.775 & 0.174 & 5.644 & 2797.034 & 0.398 & 8.750 \\
    TRUE & Random & MaskXent & 0.677 & 4.297 & 13826.641 & 0.782 & 0.176 & 3.951 & 1358.135 & 0.388 & 7.875 \\
    TRUE & Random & ValueRecon & 0.713 & 4.127 & 13668.988 & 0.777 & 0.162 & 3.760 & $\phantom{0}$986.306 & 0.396 & 6.500 \\
    TRUE & Random & MaskXent+ValueRecon & 0.701 & 4.136 & 14107.645 & 0.780 & 0.166 & 4.506 & 1917.875 & 0.397 & 8.750 \\\midrule
    FALSE & - & BinXent & 0.690 & 4.116 & \textbf{13038.762} & 0.776 & 0.170 & 3.717 & 1207.923 & 0.383 & 4.875 \\
    FALSE & - & BinRecon & 0.622 & 3.766 & 13453.309 & \textbf{0.767} & \textbf{0.158} & 3.208 & $\phantom{0}$897.645 & 0.370 & 2.250 \\
    TRUE & Const. & BinRecon & 0.634 & 3.765 & 13208.133 & 0.773 & \textbf{0.158} & \textbf{3.156} & $\phantom{0}$957.801 & 0.371 & 2.375 \\
    TRUE & Random & BinRecon & \textbf{0.619} & \textbf{3.703} & 13075.474 & 0.773 & 0.160 & 3.183 & \textbf{$\phantom{0}$870.283} & \textbf{0.368} & \textbf{1.625} \\\bottomrule          
    \end{tabular}
    }
}
\end{table*}

We first compare a series of SSL methods utilizing the same MLP encoders for each dataset.
To identify the compatibility of the binning task with other transformation functions, 
we include the cases optimizing BinRecon loss with masking. 
Finally, we experiment with four cases to validate our methodology; optimizing BinXent, treating bins as nominal classes; optimizing BinRecon, treating bins as ordinal values without any augmentation; optimizing BinRecon with masking as constant values; and optimizing BinRecon with masking as random values. 
In Table~\ref{tab:ssl}, four rows at the bottom correspond to our methods.

\paragraph{Binary classification: }
First, we compare the performance of eight datasets whose downstream task is binary classification in Table~\ref{tab:binclass}.
Interestingly, we found a consistent improvement when we changed the target for reconstruction loss from the raw values (ValueRecon) to bin indices (BinRecon) while other training details were fixed.
These results indicate that learning irregular functions (from continuous to discrete) is more beneficial than learning smooth functions (from continuous to continuous) in tabular representation learning.

\begin{table*}[tb!]
\caption{Comparison with the tree-based and deep learning methods including state-of-the-art models. 
For baselines, we directly reference the performance values from the papers to minimize ambiguity in selecting the hyperparameters. 
When the performance is not available, we leave them blank(-). 
For each dataset, the best cases among deep learning methods are marked in \textbf{bold}, and the second best results are \underline{underlined}. 
SSL+Fine-tuning methods refer to the fine-tuning results of the baseline SSL methods investigated in Section~\ref{subsec:results1}. For SSL+Fine-tuning methods corresponding to the four rows at the bottom, we provide the best results among the combinations of various input transformations (None, Masking as constant, Masking as random) and encoder networks (MLP, FT-Transformer, T2G-Former). Training details and full results are provided in Supplementary~\ref{sup_sec:full_results}.
}
\label{tab:finetuning}
\centering
\resizebox{\textwidth}{!}{
\begin{tabular}{lccccccccccccc}
\toprule
\multirow{2}{*}{Training network and method} & \multicolumn{4}{c}{Binary classification} & \multicolumn{6}{c}{Multiclass classification} & \multicolumn{3}{c}{Regression} \\\cmidrule(lr){2-5}\cmidrule(lr){6-11}\cmidrule(lr){12-14}
 & HI $\uparrow$ & PH $\uparrow$ & OS $\uparrow$ & PO $\uparrow$ & CO $\uparrow$ & GE $\uparrow$ & VO $\uparrow$ & AL $\uparrow$ & HE $\uparrow$ & MNIST $\uparrow$ & CA $\downarrow$ & HO $\downarrow$ & FI $\downarrow$ \\ \cmidrule(lr){1-1}\cmidrule(lr){2-5}\cmidrule(lr){6-11}\cmidrule(lr){12-14}
\multicolumn{13}{l}{\hspace{0.2cm}\textit{\textbf{Tree-based machine learning algorithms}}} \\
XGBoost & 0.726 & 0.721 & 0.840 & 0.711 & 0.969 & 0.683 & 0.699 & 0.924 & 0.348 & 0.977 & 0.434 & 3.152 & 10372.778 \\
CatBoost & 0.727 & 0.728 & 0.833 & 0.897 & 0.967 & 0.692 & 0.711 & 0.948 & 0.386 & 0.979 & 0.430 & 3.093 & 10636.322 \\\midrule
\multicolumn{13}{l}{\hspace{0.2cm}\textit{\textbf{Deep learning methods}}} \\
MLP & 0.714 & 0.724 & \underline{0.896} & \underline{0.901} & 0.968 & 0.659 & 0.692 & 0.960 & 0.378 & 0.983 & 0.513 & 3.146 & \underline{10086.080} \\
ResNet & 0.688 & 0.728 & 0.885 & 0.795 & 0.729 & 0.484 & 0.550 & 0.220 & 0.229 & 0.826 & 0.706 & 4.004 & 10226.508 \\
TabNet~\citep{arik2021tabnet,gorishniy2021revisiting} & 0.719 & - & - & - & 0.957 & 0.587 & 0.568 & 0.954 & 0.378 & 0.968 & 0.510 & - & - \\
NODE~\citep{popov2019node,gorishniy2021revisiting} & 0.726 & - & - & - & 0.958 & - & - & 0.918 & 0.359 & - & \underline{0.464} & - & - \\
DCN V2~\citep{wang2021dcn,gorishniy2021revisiting} & 0.723 & - & - & - & 0.965 & - & - & 0.955 & 0.385 & - & 0.484 & - & - \\
SCARF~\citep{bahri2021scarf} & 0.585 & 0.710 & 0.878 & 0.838 & 0.654 & 0.325 & 0.289 & 0.731 & 0.050 & 0.801 & 1.084 & 5.595 & 13632.255 \\
SAINT~\citep{somepalli2021saint} & 0.713 & 0.728 & 0.886 & 0.877 & 0.943 & 0.691 & 0.713 & 0.932 & 0.378 & 0.981 & 0.581 & 6.186 & 19366.582 \\  
FT-Transformer~\citep{gorishniy2021revisiting} & 0.729 & 0.724 & 0.882 & 0.890 & 0.970 & 0.664 & 0.705 & 0.960 & \textbf{0.391} & 0.966 & 0.487 & 3.319 & 10206.127 \\
PLR (MLP-Ensemble)~\citep{gorishniy2022embeddings} & \underline{0.734} & - & - & - & 0.970 & 0.674 & - & - & - & - & 0.467 & 3.050 & - \\
PLR (FT-T-Ensemble)~\citep{gorishniy2022embeddings} & \underline{0.734} & - & - & - & \textbf{0.972} & 0.646 & - & - & - & - & \underline{0.464} & 3.162 & - \\ 
T2G-Former~\citep{yan2023t2g} & \underline{0.734} & 0.746 & 0.884 & 0.881 & 0.968 & 0.656 & {0.717} & \underline{0.964} & \textbf{0.391} & \underline{0.985} & \textbf{0.455} & 3.138 & 10750.850 \\ 
SSL(MaskXent)+Fine-tuning & 0.725 & \underline{0.751} & 0.892 & 0.897 & 0.970 & \underline{0.698} & {0.717} & 0.963 & 0.383 & \underline{0.985} & 0.479 & \underline{3.086} & {10204.559} \\
SSL(ValueRecon)+Fine-tuning & 0.719 & 0.731 & 0.894 & 0.899 & 0.969 & 0.690 & 0.712 & 0.963 & 0.381 & 0.984 & 0.478 & 3.119 & 10333.400 \\
SSL(MaskXent+ValueRecon)+Fine-tuning & 0.727 & 0.737 & 0.894 & 0.896 & 0.968 & 0.658 & 0.709 & 0.959 & 0.382 & 0.984 & 0.475 & 3.257 & 10708.780 \\

\midrule\midrule
Ours -- SSL(BinRecon)+Fine-tuning & \textbf{0.737} & \textbf{0.764} & \textbf{0.897} & \textbf{0.904} & \underline{0.971} & \textbf{0.720} & \textbf{0.728}  & \textbf{0.966} & \underline{0.388} & \textbf{0.986} & \underline{0.464} & \textbf{2.989} & $\phantom{0}$\textbf{9757.950} \\
\bottomrule
\end{tabular}
}
\end{table*}

\vspace{-0.25cm}
\paragraph{Multiclass classification: }
Next, we investigate nine datasets whose downstream task is multiclass classification in Table~\ref{tab:multiclass}. 
Unlike the binary classification tasks, we observe that optimizing BinRecon loss with masking consistently leads to additional improvements compared to the cases without masking, and optimizing BinXent does not work well. 
These results indicate that the order information is important for multiclass classification and BinRecon can effectively manipulate them. Further discussion will be provided in Section~\ref{sec:discussion}.

\vspace{-0.25cm}
\paragraph{Regression: } 
Finally, we test eight datasets whose downstream task is the regression in Table~\ref{tab:regression}. 
Since the evaluation metric is RMSE, lower values correspond to better-performing cases. 
Compared to other downstream tasks, regression tasks exhibit the most significant improvements with the binning pretext task. For instance, when comparing our method with the best baselines, we observed improvements of 10.27\% for HO dataset, 8.63\% for DIA dataset, and 8.57\% for CA dataset.

\subsection{Comparison with the supervised methods: Fine-tuning results}
\label{subsec:results2}

We observed that the binning consistently improves the unsupervised learning performance across the various tabular datasets and the downstream tasks.
In this section, we compare our method against the supervised methods that utilize label information throughout the training. Our supervised baselines include tree-based algorithms, such as XGBoost~\cite{chen2016xgboost} and CatBoost~\cite{prokhorenkova2018catboost}, recent deep learning methods and fine-tuning results from SSL methods discussed in Section~\ref{subsec:results1}. Since supervised baselines often require extensive hyperparameter tuning, we directly reference the reported performances in the papers. When the performance has not been reported in the paper, we train with the default setup as depicted in the paper or leave them blank. 
For our methods, we first train encoder networks using the default setup with BinRecon loss in an unsupervised manner. Then, we conduct fine-tuning on the pre-trained encoders. The training details are provided in Supplementary~\ref{sup_sec:full_results}.

The results are summarized in Table~\ref{tab:finetuning} and Table~\ref{tab:finetuning_full}, \ref{tab:finetuning_fulldata} in the supplementary material. 
Surprisingly, our method consistently outperforms both the tree-based and deep learning methods, even though it relies solely on changing the objective function to discretized bins during pre-training.
On average, we outperform XGBoost by 5.55\% (max. 27.14\%), CatBoost by 2.18\% (max. 8.26\%), the state-of-the-art deep learning method (T2G-Former) by 2.30\% (max. 9.76\%), and the fine-tuning results of other SSL methods by 1.55\% (max. 4.38\%).

The superior performance of our method is primarily attributed to its unsupervised pretraining phase, a strategy particularly effective in deep learning and absent in tree-based algorithms. 
The key to its success lies in manipulating an appropriate inductive bias during pretraining. For our method, the binning objective effectively leverages the irregularities and mitigates the heterogeneity between the features as described in Section~\ref{sec:methods}.
Thanks to the successful implementation of this pretraining strategy, our approach achieves superior performance across a wide array of datasets.
\vspace{-0.2cm}
\section{Discussion}
\label{sec:discussion}

\subsection{Ablation study on the individual factor for binning}
\label{subsec:ablation}

In this section, we scrutinize the individual contributions of the components of binning, detailed in Section~\ref{sec:methods}.
Specifically, we examine the roles of discerning the order of samples within each feature, standardizing all features into equal sets, and grouping similar values.
BinRecon encapsulates all three elements while ValueRecon disregards them completely.
To dissect the influence of each factor, we systematically eliminate them one by one from the BinRecon loss as follows. 

\vspace{-0.4cm}
\begin{itemize}[leftmargin=*]
    \item Ordering: 
    We shuffle the bin indices with different random seeds for each feature.
    \vspace{-0.2cm}
    \item Standardizing into equal sets: 
    We replace the bin indices with the averages for each bin. Then, each feature includes different elements in different ranges. 
    \vspace{-0.2cm}
    \item Grouping: 
    We set $T^j=|\mathcal{D}^j_\text{train}|$ for every feature. In this case, each unique value corresponds to an individual bin, and only the order information remains.
\end{itemize}
\vspace{-0.3cm}

As shown in Table~\ref{tab:discussion_ablation} in supplementary material,
we found that eliminating the standardizing factor shows the largest performance degradation, averaging a 6.85\% decrease in 15 datasets among 25. 
This decline is much steeper than the effect of eliminating all three factors. From these observations, we infer that the standardizing factor which makes all features lie on the uniform distribution with identical elements is most critical for the successful implementation of binning.

\vspace{-0.2cm}
\subsection{Dependency between the number of bins and downstream task performance}

In this section, we investigate the relationship between the number of bins and downstream task performance for BinXent and BinRecon without input transformation. 
As shown in Figure~\ref{fig:ablation-numbins} in supplementary material, there is no clear relationship between the number of bins and normalized performance (Pearson correlation $\rho^2=0.01$, Kendall rank correlation $\tau=0.16$ for BinXent, $\rho^2=0.04$, $\tau=0.27$ for BinRecon), 
except that the number of bins should be not too small, but larger is not always better. 
This result is not surprising, as utilizing too few bins can eliminate necessary information while utilizing too many bins can diminish the benefits of binning.
However, a relatively strong dependency of $\rho^2=0.34$ and $\tau=0.60$ was observed in a subset of examples: regression tasks with BinRecon loss with fewer than 100 bins.

\vspace{-0.2cm}
\subsection{
Bin information is not usable unless it is provided as a pretext task
}
\label{subsec:discussion_bininfo}

So far, we found that bin information is critical for achieving superior representations across various tabular data problems.
However, even if we do not employ bin information as an explicit pretext task, it remains accessible from the raw values.
In this section, we evaluate how accurately the learned representations can predict bin indices when we optimize ValueRecon or MaskXent during SSL. 
To gauge this, we measure the relative error increase against the results of BinRecon case.
As shown in Table~\ref{tab:bin_error} in the supplementary material, the prediction error is steeply increased at an average of 66.3\% when bin information is not provided. This underscores that while bin information can be derived from the data, its utility is markedly compromised unless it is adopted as a pretext task.

\vspace{-0.2cm}
\subsection{Visualization analysis}
\label{subsec:discussion_vis}

\begin{figure}[tb]
    \centering
    \includegraphics[width=0.36\textwidth]{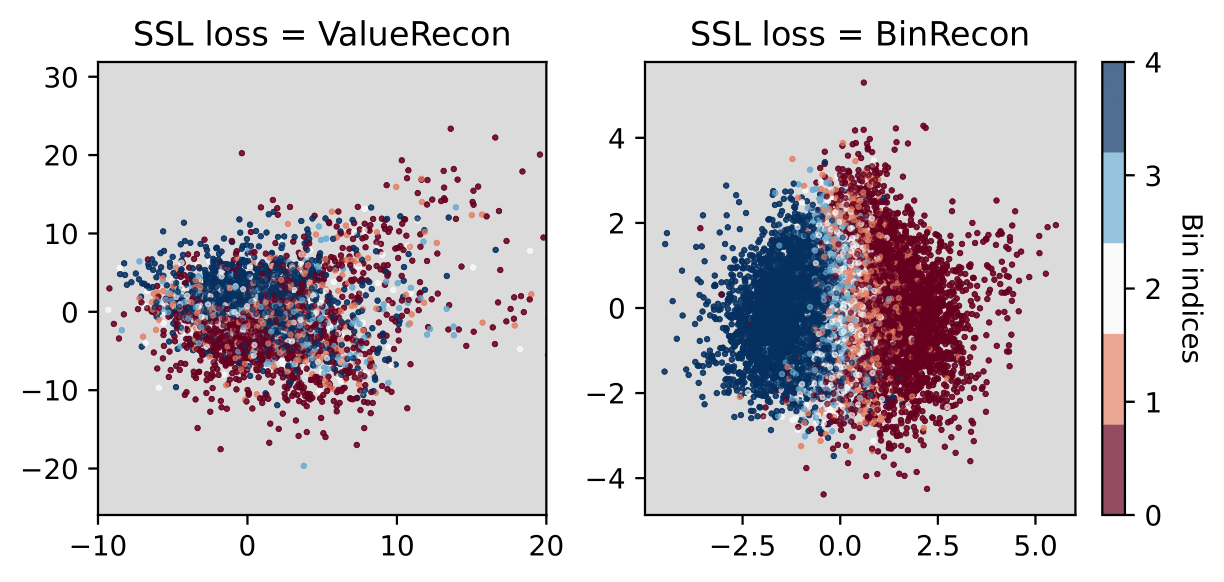}
    \vspace{-0.2cm}
    \caption{Visualization analysis using HO dataset. For better interpretability, we implement PCA for the learned representation vectors based on the different objective functions, plotting the first two principal components. Colors denote the bin indices of each sample. 
    }
    \label{fig:vis_results}
    \vspace{-0.6cm}
\end{figure}

To demonstrate the superior capability of the binning task in effectively capturing irregular functions mapping continuous inputs to discretized bin indices, compared to other methods, we present a visualization analysis of representation vectors after SSL in Figure~\ref{fig:vis_results}.
Due to the high-dimensional nature of the representation vectors, we implement PCA for better interpretability. In the visualization, the bin indices are represented as different colors. 
A distinct pattern emerges from this analysis: the representation vectors are specifically grouped according to their bin indices in the case of BinRecon. This pronounced clustering is not evident when other pretext tasks are employed. These findings highlight the effectiveness of binning as a pretext task. It demonstrates the unique capacity of this approach to enable the encoder to accurately capture the irregular function, distinguishing it from other methods.

\begin{table*}[thb!]
\caption{Comparison of fine-tuning performance for tabular SSL when applying binning as data augmentation (Randomized Quantization, RQ) versus using binning to define output labels (Ours).}
\label{tab:rq}
\centering
\resizebox{0.86\textwidth}{!}{
\begin{tabular}{lccccccccccccc}
\toprule
Training method & HI $\uparrow$ & PH $\uparrow$ & OS $\uparrow$ & PO $\uparrow$   & CO $\uparrow$   & GE $\uparrow$ & VO $\uparrow$ & AL $\uparrow$ & HE $\uparrow$   & MNIST $\uparrow$ & CA $\downarrow$ & HO $\downarrow$ & FI $\downarrow$      \\\midrule
RQ~\citep{wu2023randomized}    & 0.717 & 0.736 & 0.896 & 0.886 & 0.969 & 0.690 & 0.719 & 0.959 & 0.379 & 0.984  & 0.475 & 3.159 & 10398.616 \\
Ours & \textbf{0.737} & \textbf{0.764} & \textbf{0.897} & \textbf{0.904} & \textbf{0.971} & \textbf{0.720} & \textbf{0.728} & \textbf{0.966} & \textbf{0.388} & \textbf{0.986}  & \textbf{0.464} & \textbf{2.989} & \textbf{$\phantom{0}$9757.950} \\\bottomrule
\end{tabular}
}
\end{table*}

\subsection{Optimizing multiple loss functions during SSL}

In Section~\ref{sec:methods}, we introduce the potential for integrating multiple loss functions in SSL by employing various decoders that share a common input representation, $z$. This strategy underlines the flexibility of our approach, though it was not the primary focus of our current study.

Our preliminary investigations into the GE dataset revealed performance gains from this multi-decoder strategy. Specifically, under conditions of 2 bins and a 0.1 random masking probability, training with a single MaskXent or ValueRecon loss yielded linear probing performance of 0.509 and 0.553, individually. On the other hand, training with a single BinRecon loss yielded a linear probing performance of 0.560. Further, introducing an additional decoder to simultaneously optimize both BinRecon and MaskXent losses (with equal weights) or both BinRecon and ValueRecon losses (with equal weights) improved linear probing performance to 0.577. These observations imply that incorporating binning loss with other SSL objectives such as MaskXent could improve tabular representation learning.

\subsection{Binning as an input transformation}
\label{subsec:rq}

\citet{wu2023randomized} introduced Randomized Quantization (RQ) as a data-agnostic augmentation strategy for contrastive representation learning, which applies binning directly to the \textit{input samples}. In contrast, our method primarily utilizes classical binning on \textit{output labels} within an auto-encoding-based self-supervised learning framework.

To determine the more effective application of binning for tabular representation learning, we implemented the RQ augmentation using the official codes, following the same experimental setups as our baseline methods, as detailed in both the manuscript and supplementary materials. For the hyperparameters of the RQ augmentation, we selected the same range as our method: $\left\{2, 5, 10, 20, 50, 100\right\}$. 

As summarized in Table~\ref{tab:rq}, our approach consistently outperformed the RQ method. According to \citet{wu2023randomized}, employing binning as an augmentation strategy leads to inevitable information loss within the input samples. While in domains with inherent redundancy (e.g., images), some information loss can be mitigated through other channels or local patterns, the tabular domain typically lacks such compensatory mechanisms. For example, in a medical dataset predicting diabetes, reducing detail in a critical feature like blood sugar levels cannot be compensated by other variables due to the independence of features within tabular data. Consequently, we anticipate that employing binning as an input transformation in the tabular domain may not be effective, as it would lead to the systematic removal of vital information.
\vspace{-0.2cm}
\section{Conclusion}

In this work, we suggest a novel pretext task based on binning which can manipulate the unique properties of tabular datasets. The binning task can effectively address the challenges in tabular SSL, including mitigating the feature heterogeneity and learning the irregularities. 
Importantly, our method focuses exclusively on modifying the objective function and is independent of specific architectures or augmentation methods.
Based on the extensive experiments, we found that the binning task not only improves the unsupervised representation learning 
but also is a powerful pretraining strategy to achieve consistently superior performance against the tree-based and other deep learning methods.
In this study, we have uncovered the potential of leveraging the inherent properties of tabular data as pretext tasks for SSL. 
However, many unique characteristics remain unexplored, such as hierarchical relationships between features. We hope our work inspires further investigations into tabular-data-specific SSL in the future.

\section*{Potential broader impact}

This paper contributes to advancing the field of Machine Learning, particularly focusing on tabular data, a domain prevalent in numerous real-world applications. Our work holds the potential to significantly enhance data analysis and predictive modeling across various sectors, including healthcare, finance, and social sciences, where tabular data is extensively used. While we believe our method can lead to positive societal impacts, such as improved decision-making and more efficient data processing, we also acknowledge the importance of responsible use. It is crucial to ensure that the deployment of these advanced machine learning techniques is carried out with ethical considerations and a commitment to mitigating biases. We hope this research inspires further innovations in machine learning while prompting continuous discussion on its ethical and societal implications.

\bibliography{99_reference}
\bibliographystyle{icml2024}

\clearpage
\appendix
\onecolumn

\appendix

\section{Dataset detail}
\label{sup_sec:dataset}

In this study, we use 25 public datasets mostly from the OpenML~\citep{vanschoren2014openml} library, including the frequently used datasets in previous studies~\citep{yoon2020vime,ucar2021subtab,gorishniy2021revisiting,gorishniy2022embeddings}. 
Each dataset has exactly one train-validation-test split, so all algorithms use the same splits as the previous studies~\citep{gorishniy2021revisiting,gorishniy2022embeddings,rubachev2022revisiting}.
We summarize the main properties of datasets in Table~\ref{tab:dataset}. For each dataset, we use a predefined batch size depending on the number of training samples: 64 when the number of training samples is less than 1000, 128 when the number of training samples is larger than 1000 and less than 5000, 256 when the number of training samples is larger than 5000 and less than 10000, 512 when the number of training samples is larger than 10000 and less than 50000, and 1024 when the number of training samples is larger than 50000.

We regard the feature as categorical when the number of unique values in the training dataset is less than 20 (5 for AL, MNIST, p-MNIST, MI). The categorical variables are fed into the feature tokenizer for FT-Transformer while MLP has no additional operation for them. For MNIST and p-MNIST datasets, we ignore the features that have only one possible value throughout the training dataset.
%

In this study, we introduce a new p-MNIST dataset as a simple modification of well-known MNIST dataset. 
In constructing the p-MNIST dataset, we permute the pixel values across all samples based on a single, predefined order. Specifically, we first generate a permutation of the pixel indices ([0, 783]) using a fixed random seed. This pre-determined order is then consistently applied to the pixel values in all images within the whole dataset. The primary intention behind this methodology is to disrupt the inherent locality present in MNIST images (i.e. nearby columns are more related), thereby rendering the data more tabular-like, where spatial locality is less apparent or quantifiable (i.e. nearby columns are not necessarily more related).

\begin{table}[h]
    \centering
    \caption{Dataset summary.}
    \label{tab:dataset}
    \resizebox{\textwidth}{!}{
    \begin{tabular}{llccccccc}
    \toprule
        Abbr. & Name & \# Train & \# Validation & \# Test & \# Num & \# Cat & Task type & Batch size \\\midrule 
        CH & Churn Modeling~\tablefootnote{\url{https://www.kaggle.com/datasets/shrutimechlearn/churn-modelling}} & 
        6400 & 1600 & 2000 & 4 & 6 & Binclass & 256 \\
        HI & Higgs Small~\citep{baldi2014searching} & 62751 & 15688 & 19610 & 24 & 4 & Binclass & 1024 \\
        AD & Adult~\citep{kohavi1996scaling} & 26048 & 6513 & 16281 & 2 & 12 & Binclass & 512 \\
        BM & Bank Marketing~\citep{moro2011using} & 28934 & 7234 & 9043 & 7 & 9 & Binclass & 512 \\
        PH & Philippine~\citep{guyon2019analysis} & 3732 & 933 & 1167 & 308 & 0 & Binclass & 128 \\
        OS & Online Shoppers~\citep{sakar2019real} & 7891 & 1973 & 2466 & 8 & 9 & Binclass & 256 \\
        CS & German Credit dataset~\tablefootnote{\url{https://archive.ics.uci.edu/dataset/144/statlog+german+credit+data}} & 640 & 160 & 200 & 20 & 0 & Binclass & 64 \\
        PO & Phoneme & 3458 & 865 & 1081 & 5 & 0 & Binclass & 128 \\
        CO & Covertype~\citep{blackard1999comparative} & 371847 & 92962 & 116203 & 44 & 7 & Multiclass & 1024 \\
        OT & Otto Group Products~\tablefootnote{\url{https://www.kaggle.com/c/otto-group-product-classification-challenge/data}} & 39601 & 9901 & 12376 & 80 & 13 & Multiclass & 512 \\
        GE & Gesture Phase & 6318 & 1580 & 1975 & 32 & 0 & Multiclass & 256 \\
        VO & Volkert~\tablefootnote{\url{https://automl.chalearn.org/data}}~\cite{guyon2019analysis} & 37318 & 9330 & 11662 & 147 & 33 & Multiclass & 512 \\
        WQ & Wine Quality~\citep{cortez2009modeling} & 4157 & 1040 & 1300 & 11 & 0 & Multiclass & 128 \\
        AL & ALOI~\citep{geusebroek2005amsterdam} & 69120 & 17280 & 21600 & 124 & 4 & Multiclass & 1024 \\
        HE & Helena~\citep{guyon2019analysis} & 62752 & 15688 & 19610 & 27 & 0 & Multiclass & 512 \\
        MNIST & Handwritten Digit Images & 50000 & 10000 & 10000 & 627 & 90 & Multiclass & 512 \\
        p-MNIST & Permuted MNIST & 50000 & 10000 & 10000 & 627 & 90 & Multiclass & 512 \\
        CA & California Housing~\citep{pace1997sparse} & 13209 & 3303 & 4128 & 8 & 0 & Regression & 512 \\
        HO & House 16H~\tablefootnote{\label{urlref}\url{http://www.ncc.up.pt/~ltorgo/Regression/DataSets.html}} & 14581 & 3646 & 4557 & 16 & 0 & Regression & 512 \\
        FI & FIFA & 12273 & 3069 & 3836 & 28 & 0 & Regression & 512 \\
        MI & MSLR-WEB10K(Fold 1)~\citep{qin2013introducing} & 723412 & 235259 & 241521 & 131 & 5 & Regression & 1024 \\
        KI & Forward kinetics of an 8 link robot arm$^6$ & 5242 & 1311 & 1639 & 8 & 0 & Regression & 256 \\
        CPU & Computer Activity Databases$^6$ & 5242 & 1311 & 1639 & 8 & 0 & Regression & 256 \\
        DIA & Diamonds & 34521 & 8631 & 10788 & 9 & 0 & Regression & 512 \\
        EL & Electricity~\tablefootnote{\url{https://github.com/LeoGrin/tabular-benchmark}} & 24623 & 6156 & 7695 & 7 & 0 & Regression & 512 
        \\\bottomrule
    \end{tabular}
    }
\end{table}

\clearpage

\section{Implementation details}
\label{sup_sec:implementation}



We use the optimization strategy for SSL as follows. We do not tune any hyperparameter and the same configuration is applied to all cases. 
\begin{itemize}
    \item Optimizer: AdamW~\citep{loshchilov2017adamw}
    \item Learning rate: 1e-4
    \item Weight decay: 1e-5
    \item Epochs: 1000
    \item Learning rate scheduler: Cosine annealing scheduler~\citep{loshchilov2016sgdr,goyal2017accurate}
\end{itemize}

For the hyperparameters related to SSL, we tried $p_m\in\left\{ 0.1, 0.2, 0.3, 0.4, 0.5, 0.6, 0.7, 0.8, 0.9 \right\}$ and $T\in\left\{ 2, 5, 10, 20, 50, 100 \right\}$. When we combine the transformation function and binning methods, to reduce the hyperparameter space, we tried $p_m\in\left\{ 0.1, 0.2, 0.3 \right\}$ and $T\in\left\{ 2, 10 \right\}$ for MLPs, and $p_m\in\left\{ 0.1, 0.2 \right\}$ and $T\in\left\{ 2, 10 \right\}$ for FT-Transformers. After SSL, we evaluate the pre-trained representations with the linear head. The linear head is trained with different random seeds 10 times, and the average performance is reported.

For other state-of-the-art models, we directly reference the reported performance in the papers to reduce the ambiguity from the random seeds or the tuning details. 

\subsection{MLP}
For MLPs, we set the architecture when the validation performance is best under the supervised setup with the encoder network $f_e$ after the grid search on the depth (1, 2, 3, 4, 5) and the width (128, 256, 512, 1024). The representation size is determined as identical to the width of MLPs. The following decoder network $f_d$ is defined as symmetric with $f_e$.

For supervised learning, we use the same configuration of SSL summarized above, except that the learning rate is 0.001 and the number of epochs is 100.
We summarize the best setups for all datasets as follows.
\begin{table}[h!]
    \centering
    \caption{MLP architectures.}
    \label{tab:mlp-arch}
    \resizebox{0.8\textwidth}{!}{
    \begin{tabular}{cl|cl}
    \toprule
        Depth & Datasets & Width & Datasets \\\midrule
        1 & CH, HI, AD, BM, OS, FI, CS & 128 & CH, HI, AD, BM, OS, FI, MI, CA \\
        2 & MI, CPU, HE, OT, AL & 256 & CS, HE, KI, PH, HO \\
        3 & CA, KI, MNIST, EL & 512 & CPU, WQ, p-MNIST, DIA \\
        4 & WQ, p-MNIST, PH, HO, CO, GE, VO, PO & 1024 & CO, GE, VO, PO, MNIST, EL, OT, AL \\
        5 & DIA & & \\\bottomrule
    \end{tabular}
    }
\end{table}

\subsection{FT-Transformer}

We do not conduct any hyperparameter tuning for FT-Transformer, and we use the default setup defined in \cite{gorishniy2021revisiting} with the number of blocks as 3. For three large-scale datasets, such as MI, MNIST, and p-MNIST, we set the number of blocks as 1 because of the computational budget. For the representation size, we adopt the value found in MLP cases. For $f_d$, we use the MLP network whose architecture is the same as Table~\ref{tab:mlp-arch}.

\subsection{T2G-Former}
We do not conduct any hyperparameter tuning for FT-Transformer, and we use the default setup defined in \cite{yan2023t2g} with the number of layers as 3, the dimension of tokens as 192, the number of heads as 8, and the activation function as ReGLU. For the representation size, we adopt the value found in MLP cases. For $f_d$, we use the MLP network whose architecture is the same as Table~\ref{tab:mlp-arch}.

\subsection{Linear evaluation and Fine-tuning}

For linear evaluation, we use the same optimization configuration for SSL except for the learning rate of 0.01 for 100 epochs. For fine-tuning, we use the same setups of the supervised cases for 50 or 100 epochs.

\section{Full results}
\label{sup_sec:full_results}

Here, we present the comprehensive results from our manuscript, accompanied by standard deviations derived from 10 repetitions of the experiment.

\begin{landscape}
\begin{table}[htb!]
\caption{Full results of Table~\ref{tab:ssl}. We repeat the evaluation 10 times and the average and the standard deviations are provided.}
\label{tab:sup_vs_ours_full}
\subfloat[Binary classification]{
    \centering
    \resizebox{1.1\textwidth}{!}{
    \begin{tabular}{lllcccccccc}
    \toprule
    Masking & Masking value & Objective(s) & CH & HI & AD & BM & PH & OS & CS & PO \\\cmidrule(lr){1-3}\cmidrule(lr){4-11}
    FALSE & - & ValueRecon & 0.810$\pm$0.001 & 0.651$\pm$0.000 & 0.837$\pm$0.000 & 0.899$\pm$0.000 & 0.728$\pm$0.001 & 0.883$\pm$0.000 & 0.709$\pm$0.003 & 0.851$\pm$0.000 \\
    FALSE & RQ & ValueRecon & 0.816$\pm$0.000 & 0.654$\pm$0.000 & 0.842$\pm$0.000 & 0.898$\pm$0.000 & 0.727$\pm$0.001 & 0.882$\pm$0.001 & 0.725$\pm$0.002 & 0.842$\pm$0.000 \\
    TRUE & Const. & MaskXent & 0.807$\pm$0.001 & 0.672$\pm$0.000 & 0.836$\pm$0.000 & 0.899$\pm$0.000 & 0.715$\pm$0.000 & 0.893$\pm$0.000 & 0.708$\pm$0.004 & 0.845$\pm$0.000 \\
    TRUE & Const. & ValueRecon & 0.810$\pm$0.000 & 0.653$\pm$0.000 & 0.839$\pm$0.000 & 0.900$\pm$0.000 & 0.734$\pm$0.001 & 0.884$\pm$0.000 & 0.718$\pm$0.002  & 0.849$\pm$0.001 \\
    TRUE & Const. & MaskXent+ValueRecon & 0.817$\pm$0.001 & 0.669$\pm$0.000 & 0.835$\pm$0.000 & 0.900$\pm$0.000 & 0.724$\pm$0.001 & 0.877$\pm$0.000 & 0.706$\pm$0.000 & 0.837$\pm$0.002 \\
    TRUE & Random & MaskXent & 0.814$\pm$0.000 & 0.681$\pm$0.000 & 0.843$\pm$0.000 & \textbf{0.901}$\pm$0.000 & 0.710$\pm$0.000 & 0.883$\pm$0.000 & 0.706$\pm$0.000 & 0.853$\pm$0.000 \\
    TRUE & Random & ValueRecon & 0.811$\pm$0.000 & 0.661$\pm$0.000 & 0.838$\pm$0.000 & 0.898$\pm$0.000 & 0.736$\pm$0.001 & 0.885$\pm$0.000 & 0.714$\pm$0.003 & 0.842$\pm$0.000 \\
    TRUE & Random & MaskXent+ValueRecon & 0.804$\pm$0.001 & 0.647$\pm$0.000 & 0.826$\pm$0.000 & 0.899$\pm$0.000 & 0.715$\pm$0.003 & 0.879$\pm$0.001 & 0.713$\pm$0.003 & 0.861$\pm$0.001 \\\midrule
    FALSE & - & BinXent & 0.817$\pm$0.001 & 0.683$\pm$0.000 & 0.845$\pm$0.000 & \textbf{0.901}$\pm$0.000 & 0.732$\pm$0.001 & 0.886$\pm$0.000 & \textbf{0.738}$\pm$0.000 & 0.851$\pm$0.001 \\
    
    FALSE & - & BinRecon & \textbf{0.823}$\pm$0.000 & \textbf{0.687}$\pm$0.000 & 0.840$\pm$0.000 & 0.900$\pm$0.000 & \textbf{0.737}$\pm$0.000 & 0.889$\pm$0.000 & 0.724$\pm$0.005 & \textbf{0.865}$\pm$0.000 \\
    TRUE & Const. & BinRecon & 0.820$\pm$0.000 & 0.672$\pm$0.000 & 0.843$\pm$0.000 & 0.899$\pm$0.000 & 0.730$\pm$0.000 & \textbf{0.896}$\pm$0.000 & 0.718$\pm$0.004 & 0.858$\pm$0.000 \\
    TRUE & Random & BinRecon & 0.819$\pm$0.001 & 0.682$\pm$0.000 & \textbf{0.846}$\pm$0.000 & 0.898$\pm$0.000 & 0.735$\pm$0.000 & 0.894$\pm$0.000 & 0.718$\pm$0.004 & 0.858$\pm$0.000 \\\bottomrule          
    \end{tabular}
    }
    }\\
    \subfloat[Multiclass classification]{
    \resizebox{1.1\textwidth}{!}{
    \begin{tabular}{lllccccccccc}
    \toprule
    Masking & Masking value & Objective(s) & CO & OT & GE & VO & WQ & AL & HE & MNIST & p-MNIST \\\cmidrule(lr){1-3}\cmidrule(lr){4-12}
    FALSE & - & ValueRecon & 0.769$\pm$0.000 & 0.776$\pm$0.000 & 0.527$\pm$0.001 & 0.619$\pm$0.000 & 0.568$\pm$0.001 & 0.931$\pm$0.000 & 0.353$\pm$0.000 & 0.965$\pm$0.000 & 0.928$\pm$0.000 \\
    FALSE & RQ & ValueRecon & 0.775$\pm$0.000 & 0.771$\pm$0.000 & 0.531$\pm$0.001 & 0.620$\pm$0.000 & 0.559$\pm$0.000 & 0.930$\pm$0.000 & 0.350$\pm$0.000 & 0.948$\pm$0.000 & 0.706$\pm$0.000 \\
    TRUE & Const. & MaskXent & 0.784$\pm$0.000 & 0.777$\pm$0.000 & 0.518$\pm$0.001 & 0.545$\pm$0.000 & 0.547$\pm$0.000 & 0.909$\pm$0.001 & 0.341$\pm$0.000 & 0.793$\pm$0.000 & 0.554$\pm$0.000 \\
    TRUE & Const. & ValueRecon & 0.783$\pm$0.000 & 0.791$\pm$0.000 & 0.557$\pm$0.001 & 0.622$\pm$0.000 & 0.586$\pm$0.001 & 0.931$\pm$0.000 & 0.354$\pm$0.000 & 0.966$\pm$0.000 & 0.925$\pm$0.000 \\
    TRUE & Const. & MaskXent+ValueRecon & 0.750$\pm$0.000 & 0.774$\pm$0.001 & 0.519$\pm$0.005 & 0.610$\pm$0.000 & 0.571$\pm$0.001 & 0.931$\pm$0.000 & 0.360$\pm$0.000 & 0.941$\pm$0.000 & 0.907$\pm$0.000 \\
    TRUE & Random & MaskXent & 0.763$\pm$0.000 & 0.791$\pm$0.000 & 0.555$\pm$0.000 & 0.549$\pm$0.000 & 0.544$\pm$0.001 & 0.925$\pm$0.000 & 0.336$\pm$0.000 & 0.945$\pm$0.000 & 0.817$\pm$0.000 \\
    TRUE & Random & ValueRecon & 0.761$\pm$0.000 & 0.782$\pm$0.000 & 0.538$\pm$0.001 & 0.625$\pm$0.000 & 0.573$\pm$0.001
     & 0.930$\pm$0.000 & 0.357$\pm$0.000 & 0.956$\pm$0.000 & 0.934$\pm$0.000 \\
    TRUE & Random & MaskXent+ValueRecon & 0.769$\pm$0.000 & 0.779$\pm$0.001 & 0.521$\pm$0.004 & 0.564$\pm$0.001 & 0.519$\pm$0.004 & 0.925$\pm$0.001 & 0.353$\pm$0.001 & 0.945$\pm$0.000 & 0.906$\pm$0.001 \\\midrule
    FALSE & - & BinXent & 0.742$\pm$0.000 & 0.781$\pm$0.000 & 0.517$\pm$0.001 & 0.600$\pm$0.001 & 0.565$\pm$0.001 & 0.903$\pm$0.001 & 0.354$\pm$0.000 & 0.956$\pm$0.000 & 0.908$\pm$0.000 \\
    FALSE & - & BinRecon & 0.784$\pm$0.000 & 0.783$\pm$0.000 & 0.544$\pm$0.001 & 0.625$\pm$0.000 & \textbf{0.592}$\pm$0.001 & 0.935$\pm$0.000 & 0.357$\pm$0.000 & 0.964$\pm$0.000 & 0.950$\pm$0.000 \\
    TRUE & Const. & BinRecon & 0.812$\pm$0.000 & 0.792$\pm$0.000 & 0.559$\pm$0.001 & 0.647$\pm$0.000 & 0.581$\pm$0.001 & 0.943$\pm$0.000 & 0.359$\pm$0.000 & 0.974$\pm$0.000 & 0.964$\pm$0.000 \\
    TRUE & Random & BinRecon & \textbf{0.814}$\pm$0.000 & \textbf{0.794}$\pm$0.000 & \textbf{0.580}$\pm$0.000 & \textbf{0.655}$\pm$0.000 & 0.574$\pm$0.001 & \textbf{0.949}$\pm$0.000 & \textbf{0.365}$\pm$0.000 & \textbf{0.981}$\pm$0.000 & \textbf{0.971}$\pm$0.000 \\\bottomrule          
    \end{tabular}
    }
}\\
\subfloat[Regression]{
    \centering
    \resizebox{1.1\textwidth}{!}{
    \begin{tabular}{lllcccccccc}
    \toprule
    Masking & Masking value & Objective(s) & CA & HO & FI & MI & KI & CPU & DIA & EL \\\cmidrule(lr){1-3}\cmidrule(lr){4-11}
    FALSE & - & ValueRecon & 0.749$\pm$0.000 & 4.241$\pm$0.001 & 13900.720$\pm\phantom{0}$0.816 & 0.784$\pm$0.000 & 0.163$\pm$0.000 & 3.876$\pm$0.002 & 1016.641$\pm$0.191 & 0.399$\pm$0.001 \\
    FALSE & RQ & ValueRecon & 0.714$\pm$0.000 & 4.165$\pm$0.000 & 13684.367$\pm$0.778 & 0.784$\pm$0.000 & 0.162$\pm$0.000 & 3.751$\pm$0.002 & 1056.453$\pm$0.229 & 0.398$\pm$0.000 \\
    TRUE & Const. & MaskXent & 0.709$\pm$0.000 & 4.548$\pm$0.000 & 13473.750$\pm\phantom{0}$1.371 & 0.788$\pm$0.000 & 0.185$\pm$0.000 & 4.475$\pm$0.033 & 1259.744$\pm$1.066 & 0.396$\pm$0.000 \\
    TRUE & Const. & ValueRecon & 0.693$\pm$0.000 & 4.086$\pm$0.000 & 13518.683$\pm\phantom{0}$0.936 & 0.778$\pm$0.000 & 0.160$\pm$0.000 & 3.728$\pm$0.003 & 952.444$\pm$0.130 & 0.394$\pm$0.000 \\
    TRUE & Const. & MaskXent+ValueRecon & 0.700$\pm$0.001 & 4.157$\pm$0.045 & 13915.875$\pm$18.078 & 0.775$\pm$0.000 & 0.174$\pm$0.001 & 5.644$\pm$0.078 & 2797.034$\pm$191.324 & 0.398$\pm$0.001 \\
    TRUE & Random & MaskXent & 0.677$\pm$0.000 & 4.297$\pm$0.000 & 13826.641$\pm\phantom{0}$0.624 & 0.782$\pm$0.000 & 0.176$\pm$0.000 & 3.951$\pm$0.001 & 1358.135$\pm$0.191 & 0.388$\pm$0.000 \\
    TRUE & Random & ValueRecon & 0.713$\pm$0.000 & 4.127$\pm$0.000 & 13668.988$\pm\phantom{0}$1.262 & 0.777$\pm$0.000 & 0.162$\pm$0.000 & 3.760$\pm$0.002 & 986.306$\pm$0.359 & 0.396$\pm$0.000 \\
    TRUE & Random & MaskXent+ValueRecon & 0.701$\pm$0.003 & 4.136$\pm$0.028 & 14107.645$\pm$29.125 & 0.780$\pm$0.001 & 0.166$\pm$0.001 & 4.506$\pm$0.034 & 1917.875$\pm$123.359 & 0.397$\pm$0.008 \\\midrule
    FALSE & - & BinXent & 0.690$\pm$0.000 & 4.116$\pm$0.001 & \textbf{13038.762}$\pm\phantom{0}$1.618 & 0.776$\pm$0.000 & 0.170$\pm$0.000 & 3.717$\pm$0.006 & 1207.923$\pm$2.552 & 0.383$\pm$0.000 \\
    FALSE & - & BinRecon & 0.622$\pm$0.000 & 3.766$\pm$0.000 & 13453.309$\pm\phantom{0}$0.832 & \textbf{0.767}$\pm$0.000 & \textbf{0.158}$\pm$0.000 & 3.208$\pm$0.001 & 897.645$\pm$0.182 & 0.370$\pm$0.000 \\
    TRUE & Const. & BinRecon & 0.634$\pm$0.000 & 3.765$\pm$0.000 & 13208.133$\pm\phantom{0}$1.960 & 0.773$\pm$0.000 & \textbf{0.158}$\pm$0.000 & \textbf{3.156}$\pm$0.002 & 957.801$\pm$0.070 & 0.371$\pm$0.000 \\
    TRUE & Random & BinRecon & \textbf{0.619}$\pm$0.000 & \textbf{3.703}$\pm$0.000 & 13075.474$\pm\phantom{0}$0.800 & 0.773$\pm$0.000 & 0.160$\pm$0.000 & 3.183$\pm$0.001 & \textbf{870.283}$\pm$0.093 & \textbf{0.368}$\pm$0.000 \\\bottomrule          
\end{tabular}
}}
\end{table}

\end{landscape}


\begin{table}[htb!]
\caption{
Fine-tuning results with standard deviation based on 10 times repeated experiments.
In this case, the models are pretrained in an unsupervised fashion (i.e. optimizing BinRecon loss) and fine-tuned in a supervised fashion. 
For each dataset and encoder, we experiment with various combinations of input transformation methods and the number of bins as explained in Supplementary~\ref{sup_sec:implementation}. Then, we repeated the best case determined based on the validation performance. 
}
\label{tab:finetuning_full}
\centering
\resizebox{\textwidth}{!}{
\vspace{-0.5cm}
\begin{tabular}{lccccccccccccc}
\toprule
Encoder & HI $\uparrow$ & PH $\uparrow$ & OS $\uparrow$ & PO $\uparrow$ & CO $\uparrow$ & GE $\uparrow$ & VO $\uparrow$ & AL $\uparrow$ & HE $\uparrow$ & MNIST $\uparrow$ & CA $\downarrow$ & HO $\downarrow$ & FI $\downarrow$ \\\midrule
MLP & 0.717$\pm$0.001 & 0.738$\pm$0.009 & \textbf{0.897$\pm$0.000} & 0.893$\pm$0.004 & 0.969$\pm$0.000 & 0.673$\pm$0.004 & \textbf{0.728$\pm$0.001} & 0.963$\pm$0.001 & \textbf{0.388$\pm$0.001} & \textbf{0.986$\pm$0.000} & 0.502$\pm$0.002 & \textbf{2.989$\pm$0.015} & $\phantom{0}$9963.609$\pm\phantom{0}$23.173 \\ 
FT-Transformer & 0.703$\pm$0.004 & 0.742$\pm$0.011 & 0.882$\pm$0.004 & \textbf{0.904$\pm$0.003} & \textbf{0.971$\pm$0.000} & 0.698$\pm$0.006 & 0.720$\pm$0.003 & 0.961$\pm$0.001 & 0.374$\pm$0.002 & 0.978$\pm$0.001 & 0.475$\pm$0.003 & 3.173$\pm$0.024 & \textbf{$\phantom{0}$9757.950$\pm$210.751} \\ 
T2G-Former & \textbf{0.737$\pm$0.001} & \textbf{0.764$\pm$0.008} & 0.892$\pm$0.003 & 0.895$\pm$0.005 & 0.967$\pm$0.001 & \textbf{0.720$\pm$0.002} & 0.725$\pm$0.001 & \textbf{0.966$\pm$0.001} & 0.378$\pm$0.002 & 0.985$\pm$0.000 & \textbf{0.464$\pm$0.001} & 3.144$\pm$0.041 & 10155.818$\pm$132.559 \\
\bottomrule
\end{tabular}
}
\vspace{-0.2cm}
\end{table}

\begin{table}[htb!]
\caption{
Fine-tuning results using a fixed set of hyperparameters across all datasets (encoder network: T2G-Former, input transformation: random masking with a 0.2 ratio, and the number of bins=10). 
In this case, the models are pretrained in an unsupervised fashion (i.e. optimizing BinRecon loss) and fine-tuned in a supervised fashion. 
Even with these fixed parameters, our method maintains a competitive edge over baseline approaches, thereby affirming the intrinsic strength and adaptability of our approach. Specifically, under this setting, our method still showcases notable performance improvements across a range of datasets, reaffirming its effectiveness beyond the scope of extensive hyperparameter optimization.
}
\label{tab:finetuning_fixed}
\centering
\resizebox{\textwidth}{!}{
\vspace{-0.5cm}
\begin{tabular}{lcccccccccccccc}
\toprule
Encoder & HI $\uparrow$ & PH $\uparrow$ & OS $\uparrow$ & PO $\uparrow$ & CO $\uparrow$ & GE $\uparrow$ & VO $\uparrow$ & AL $\uparrow$ & HE $\uparrow$ & MNIST $\uparrow$ & CA $\downarrow$ & HO $\downarrow$ & FI $\downarrow$ & Average Rank\\\midrule
XGBoost & 0.726 & 0.721 & 0.840 & 0.711 & 0.969 & 0.683 & 0.699 & 0.924 & 0.348 & 0.977 & 0.434 & 3.152 & 10372.778 3.462 \\ 
SSL+Finetuning(T2G-Former, Random masking(0.2), MaskXent) & 0.714 & 0.733 & 0.872 & 0.895 & 0.925 & 0.708 & 0.705 & 0.960 & 0.365 & 0.983 & 0.551 & 3.174 & 10201.881 & 2.692 \\
SSL+Finetuning(T2G-Former, Random masking(0.2), ValueRecon) & 0.719 & 0.727 & 0.870 & 0.892 & 0.874 & 0.673 & 0.713 & 0.762 & 0.343 & 0.982 & 0.474 & 3.310 & 10434.967 & 4.000 \\
SSL+Finetuning(T2G-Former, Random masking(0.2), MaskXent+ValueRecon) & 0.721 & 0.757 & 0.873 & 0.891 & 0.839 & 0.657 & 0.699 & 0.958 & 0.351 & 0.983 & 0.478 & 3.101 & 10063.307 & 3.000 \\
Ours(T2G-Former, Random masking(0.2), Bin=10, BinRecon) & 0.734 & 0.773 & 0.880 & 0.895 & 0.965 & 0.699 & 0.728 & 0.963 & 0.380 & 0.983 & 0.469 & 3.193 & 10006.578 & 1.462 \\
\bottomrule
\end{tabular}
}
\vspace{-0.2cm}
\end{table}

We also summarize the detailed training setups for the best cases in Table~\ref{tab:finetuning_full} as follows.

\begin{table}[htb!]
\caption{
Training setups for the best cases in Table~\ref{tab:finetuning_full}.
}
\centering
\resizebox{\textwidth}{!}{
\begin{tabular}{lccccccccccccc}
\toprule
Datasets & HI & PH & OS & PO & CO & GE & VO & AL & HE & MNIST & CA & HO & FI \\\midrule
Encoders & T2G-Former & T2G-Former & MLP & FT-Transformer & FT-Transformer & T2G-Former & MLP & T2G-Former & MLP & MLP & T2G-Former & MLP & FT-Transformer \\
Input transformation & Masking(Random) & Masking(Random) & None & Masking(Random) & Masking(Const.) & Masking(Random) & Masking(Const.) & Masking(Random) & Masking(Random) & Masking(Random) & Masking(Const.) & Masking(Random) & Masking(Const.) \\
Masking probability ($p_m$) & 0.1 & 0.2 & - & 0.1 & 0.2 & 0.2 & 0.3 & 0.2 & 0.2 & 0.3 & 0.2 & 0.2 & 0.2 \\
Number of bins & 2 & 10 & 2 & 10 & 10 & 10 & 10 & 10 & 2 & 10 & 10 & 2 & 2 \\
Fine-tuning epochs & 50 & 50 & 50 & 50 & 100 & 100 & 100 & 100 & 100 & 100 & 50 & 100 & 100 \\
\bottomrule
\end{tabular}
}
\end{table}

Here are the results for the other list of datasets, not included in Table~\ref{tab:finetuning}. Again, we found that the binning method consistently outperforms other methods.

\begin{table*}[h!]
\caption{Fine-tuning results for the other list of datasets, not included in Table~\ref{tab:finetuning}.}
\label{tab:finetuning_fulldata}
\centering
\resizebox{0.85\textwidth}{!}{
\begin{tabular}{lcccccccccccc}
\toprule
\multirow{2}{*}{Training network and method} & \multicolumn{4}{c}{Binary classification} & \multicolumn{3}{c}{Multiclass classification} & \multicolumn{5}{c}{Regression} \\\cmidrule(lr){2-5}\cmidrule(lr){6-8}\cmidrule(lr){9-13}
 & CH $\uparrow$ & AD $\uparrow$ & BM $\uparrow$ & CS $\uparrow$ & OT $\uparrow$ & WQ $\uparrow$ & p-MNIST $\uparrow$ & MI $\downarrow$ & KI $\downarrow$ & CPU $\downarrow$ & DIA $\downarrow$ & EL $\downarrow$ \\ \cmidrule(lr){1-1}\cmidrule(lr){2-5}\cmidrule(lr){6-8}\cmidrule(lr){9-13}
\multicolumn{12}{l}{\hspace{0.2cm}\textit{\textbf{Tree-based machine learning algorithms}}} \\
XGBoost & 0.859 & 0.875 & 0.903 & 0.710 & 0.827 & 0.632 & 0.978 & 0.742 & 0.128 & 15.437 & 564.547 & 0.293 \\
CatBoost & 0.861 & 0.873 & 0.907 & 0.750 & 0.825 & 0.659 & 0.980 & 0.741 & 0.090 & 2.668 & 531.584 & 0.291 \\\midrule
\multicolumn{12}{l}{\hspace{0.2cm}\textit{\textbf{Deep learning methods}}} \\
MLP & 0.838 & 0.851 & 0.902 & 0.666 & 0.810 & 0.629 & \underline{0.980} & 0.753 & 0.072 & 2.764 & 563.123 & 0.354 \\ 
ResNet & 0.827 & 0.842 & 0.903 & \underline{0.750} & 0.745 & 0.570 & 0.806 & 0.769 & 0.160 & 3.517 & 919.240 & 0.409 \\
FT-Transformer~\citep{gorishniy2021revisiting} & 0.831 & 0.836 & \underline{0.904} & 0.676 & 0.796 & 0.618 & 0.957 & \textbf{0.746} & 0.073 & 2.746 & \underline{538.575} & 0.350 \\
T2G-Former~\citep{yan2023t2g} & \textbf{0.863} & \textbf{0.860} & 0.903 & 0.681 & \textbf{0.819} & 0.599 & \underline{0.980} & 0.754 & \underline{0.069} & 2.708 & 544.061 & 0.350 \\ 
SSL(RQ)+Fine-tuning & \underline{0.843} & 0.852 & 0.900 & 0.695 & 0.812 & 0.627 & 0.978 & 0.757 & 0.073 & 2.709 & 530.547 & 0.346 \\
SSL(MaskXent)+Fine-tuning & 0.841 & 0.849 & \underline{0.904} & 0.713 & {0.816} & \underline{0.639} & 0.978 & 0.753 & 0.071 & 2.786 & 538.677 & \textbf{0.338} \\
SSL(ValueRecon)+Fine-tuning & 0.837 & 0.851 & \underline{0.904} & 0.681 & {0.816} & 0.631 & 0.978 & 0.753 & 0.072 & \underline{2.688} & 541.866 & 0.347  \\
SSL(MaskXent+ValueRecon)+Fine-tuning & 0.836 & 0.848 & 0.902 & 0.725 & 0.815 & 0.628 & 0.978 & 0.752 & 0.071 & 2.712 & 576.607 & 0.344  \\
\midrule\midrule
Ours -- SSL(BinRecon)+Linear eval/Fine-tuning & \underline{0.843} & \underline{0.857} & \textbf{0.910} & \textbf{0.774} & \underline{0.817} & \textbf{0.648} & \textbf{0.982} & \underline{0.750} & \textbf{0.068} & \textbf{2.686} & \textbf{531.458} & \underline{0.339} \\
\bottomrule
\end{tabular}
}
\end{table*}

\clearpage

\section{Additional results for discussion}

\subsection{Comparing the binning method between the quantiles and the equal-width}
\label{sup_sec:binmethod}

We found that the grouping is critical for implementing the binning task successfully. Instead of quantile-based binning in our method, we also can manipulate equal-width binning. Here, we experiment with which method can be more beneficial for binning between the quantile and fixed size. We test the same candidates for the number of bins for equal-width binning, and we compare the test performance when the validation performance is the best with the quantile-based ones. 

\begin{figure}[h!]
    \centering
    \includegraphics[width=0.35\textwidth]{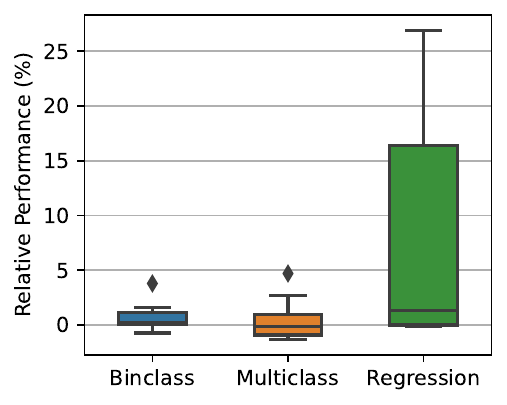}
    \caption{Relative performance when we change the binning method to the equal-width from the quantiles. When the values are positive, the quantile-based binning is better than the equal-width binning. When the values are negative, vice versa. In particular, for regression tasks, the quantile-based binning is much better than the equal-width binning.
    }
    \label{fig:discussion-binmethod}
\end{figure}

The results are described in Figure~\ref{fig:discussion-binmethod}. Among 25 datasets, equal-width binning showed better performance for three datasets (PH, HE, MNIST) to the extent of 0.6\% at the maximum, and two binning methods showed comparable performance for two datasets (OT, AL). For the other 20 datasets, quantile-based binning showed better performance. In particular, for regression tasks, we found that the performance degrades 27\% as the maximum when we change the binning method from quantile to fixed size. Finally, we conclude that quantile-based binning consistently results in good representations across various datasets. 

\subsection{Dependency between the number of bins and downstream task performance}

We investigate the relationship between the number of bins and downstream task performance for BinXent and BinRecon without input transformation. 
Because the performance range is quite different between the datasets, we normalize the performance with the best and worst cases for each dataset. 
This approach allows us to normalize performance metrics across datasets with varying ranges and different evaluation metrics. Specifically, we assess the best and worst performance among six models that differ by the number of bins (2, 5, 10, 20, 50, 100), keeping the loss function (BinXent, BinRecon) consistent. Consequently, the normalized scale sets the best-performing case to 1 and the worst-performing case to 0. 

\begin{figure}[h]
    \centering
    \includegraphics[width=0.27\textwidth]{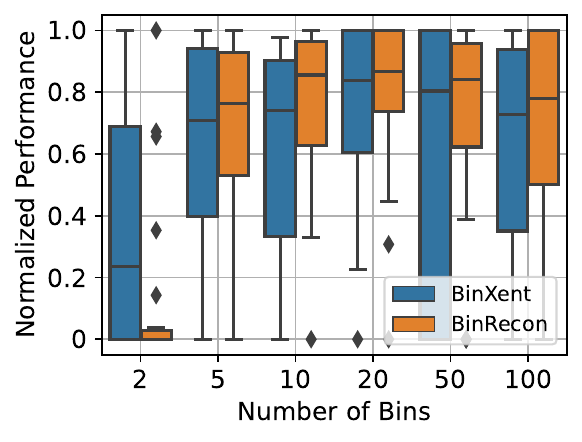}
    \caption{Empirical analysis on the dependency between the number of bins and the downstream task performance. 
    }
    \label{fig:ablation-numbins}
\end{figure}

\clearpage

\subsection{Results for Section~\ref{sec:discussion}}

\begin{table}[h!]
    \centering
    \caption{
    Ablation test results on individual components of binning.
    }
    \label{tab:discussion_ablation}
    \resizebox{0.42\textwidth}{!}{
    \begin{tabular}{ccccc}
    \toprule
    Ordering & Standardizing & Grouping & Improved & Deteriorated \\\midrule
    Yes & Yes & Yes & - (\textit{Baseline}) & - (\textit{Baseline}) \\
    No & Yes & Yes & 1 ($+$4.70\%) & 12 ($-$4.05\%) \\
    Yes & No & Yes & 1 ($+$5.21\%) & 15 ($-$6.85\%) \\
    Yes & No & No & 3 ($+$1.95\%) & 12 ($-$5.83\%) \\
    No & No & No & - & 18 ($-$6.02\%)  \\\bottomrule
    \end{tabular}
    }
\end{table}

\begin{table}[htb!]
\centering
\caption{
Binning regression task performance on various SSL methods. We provide the relative error with the baseline of the BinRecon case. For all cases, the error is increased by at least 38\%. As a result, the binning indices are achievable from the raw inputs but not usable in the resulting representations when we do not explicitly provide as the pretext targets.
}
\label{tab:bin_error}
\resizebox{0.6\textwidth}{!}{
\begin{tabular}{lllr}
\toprule
Masking & Masking value & Objective(s) & Relative error increase (\%)
\\\midrule\midrule
False & - & BinRecon & \textit{(Baseline)} 0 \\\midrule
False & - & ValueRecon & $\phantom{0}$49.579 \\
True & Const. & MaskXent & $\phantom{0}$82.922 \\
True & Const. & ValueRecon & $\phantom{0}$38.444 \\
True & Const. & MaskXent+ValueRecon & $\phantom{0}$68.344 \\
True & Random & MaskXent & 111.708 \\
True & Random & ValueRecon & $\phantom{0}$38.135 \\
True & Random & MaskXent+ValueRecon & $\phantom{0}$60.016
 \\\midrule
\multicolumn{3}{c}{Average} & $\phantom{0}$66.285 \\\bottomrule
\end{tabular}
}
\end{table}

\subsection{Additional results for Section~\ref{subsec:rq}}

\begin{table*}[thb!]
\caption{Comparison of fine-tuning performance for tabular SSL when applying binning as data augmentation (Randomized Quantization, RQ) versus using binning to define output labels (Ours).}
\centering
\resizebox{\textwidth}{!}{
\begin{tabular}{lccccccccccccc}
\toprule
Training method & HI $\uparrow$ & PH $\uparrow$ & OS $\uparrow$ & PO $\uparrow$   & CO $\uparrow$   & GE $\uparrow$ & VO $\uparrow$ & AL $\uparrow$ & HE $\uparrow$   & MNIST $\uparrow$ & CA $\downarrow$ & HO $\downarrow$ & FI $\downarrow$      \\\midrule
RQ~\citep{wu2023randomized}    & 00.717$\pm$0.002 & 0.736$\pm$0.005 & 0.896$\pm$0.002 & 0.886$\pm$0.004 & 0.969$\pm$0.000 & 0.690$\pm$0.007 & 0.719$\pm$0.003 & 0.959$\pm$0.000 & 0.379$\pm$0.001 & 0.984$\pm$0.001  & 0.475$\pm$0.002 & 3.159$\pm$0.030 & 10398.616$\pm\phantom{0}$28.659 \\
Ours & 0.737$\pm$0.001 & 0.764$\pm$0.008 & 0.897$\pm$0.000 & 0.904$\pm$0.003 & 0.971$\pm$0.000 & 0.720$\pm$0.002 & 0.728$\pm$0.001 & 0.966$\pm$0.001 & 0.388$\pm$0.001 & 0.986$\pm$0.000  & 0.464$\pm$0.001 & 2.989$\pm$0.015 & $\phantom{0}$9757.950$\pm$210.751 \\\bottomrule
\end{tabular}
}
\end{table*}

\section{Additional descriptions}

\subsection{Impact on uninformative features}
\label{supsec:uninformative}

\citet{muller2021transformers} and \citet{stewart2023regression} demonstrated that incorporating discretization into the objective function of deep networks not only improves training efficiency but also proves to be a theoretically sound method for modeling any distribution. This underlines the significant potential of binning to enhance neural network performance. We propose that one key advantage of binning is its ability to simplify the dataset into $T$ distinct values per feature, creating equal sets among all features. This property helps prevent uninformative features—those with low mutual information with the task label but potentially high entropy due to variance or a large number of unique values—from dominantly influencing the training process.

In scenarios where tabular SSL is applied using straightforward reconstruction loss, neural networks might inadvertently focus more on features characterized by their high variability(frequency) or unique value counts. This phenomenon, more pronounced in tabular data as noted in recent studies~\citep{beyazit2023inductive,cherepanova2024performance}, suggests that training could be skewed towards these high-frequency yet less informative features. By substituting output labels with bin indices during SSL, our method explicitly constrain all the features to regress on the SSL outputs with same frequency or same variability. Thus, it effectively circumvents that any specific feature from dominating during SSL, ensuring that such uninformative features do not overshadow the learning of meaningful representations.

\end{document}